\documentclass[12pt]{clear2026} 

\jmlrproceedings{}{}
\jmlrvolume{}
\jmlryear{}
\jmlrworkshop{}

\title[Causal Parametric Drift Simulation]{Causal Parametric Drift Simulation: A Digital Twin Framework for Classifier Robustness Evaluation}
\usepackage{times}
\usepackage{amsmath}
\usepackage{booktabs}
\usepackage{multirow}
\usepackage{tikz}
\usepackage{float}
\usepackage{pifont}
\usepackage{colortbl}

\AtBeginDocument{%
  \hypersetup{%
    pdftitle={Causal Parametric Drift Simulation: A Digital Twin Framework for Classifier Robustness Evaluation},%
    pdfauthor={Julien Lafrance, Richard Khoury, V\'eronique Tremblay},%
    pdfsubject={Concept drift, Structural Causal Models, Digital Twins, classifier robustness},%
    pdfkeywords={Concept Drift; Structural Causal Models; Digital Twins; Synthetic Data; Robustness; Causal Discovery}%
  }%
}
\newcommand{\cmark}{\ding{51}}
\newcommand{\xmark}{\ding{55}}
\usetikzlibrary{positioning, shapes}

\clearauthor{%
 \Name{Julien Lafrance} \Email{julien.lafrance.1@ulaval.ca}\\
 \addr Laval University
 \AND
 \Name{Richard Khoury} \Email{richard.khoury@ift.ulaval.ca}\\
 \addr Laval University
 \AND
 \Name{V\'eronique Tremblay} \Email{Veronique.Tremblay@mat.ulaval.ca}\\
 \addr Laval University
}
\begin{document}

\maketitle

\begin{abstract}
  Machine learning classifiers in dynamic environments face concept drift---changes in the data-generating process that degrade performance. Conventional evaluation via static test sets or noise perturbations fails to preserve causal dependencies in tabular data, often producing causally invalid assessments. Post-hoc tools like SHAP and LIME offer correlational insights that may not reflect the causal mechanisms driving model failure.

  We propose a framework that complements existing drift detection by leveraging Structural Causal Models as ``Digital Twins'' of data-generating processes, enabling precise causal interventions while preserving structural dependencies. Our technique, Causal Parametric Drift Simulation, stress-tests classifiers to identify vulnerabilities before deployment. Experiments on the Open Sourcing Mental Illness (OSMI) dataset demonstrate that this approach exposes latent vulnerabilities invisible to standard statistical monitors. Code and experimental scripts are publicly available.
\end{abstract}


\section{Introduction}

Machine learning classification models deployed in dynamic environments face the persistent challenge of \emph{concept drift}: a shift in the relationship between features and the classification target \citep{hashmani_concept_2020}. Left unaddressed, concept drift gradually erodes predictive performance, as seen when COVID-19 invalidated pre-pandemic diagnostic models or when market regime changes degraded credit scoring systems. This makes concept drift a critical failure mode in high-stakes domains such as healthcare and finance \citep{m_s_survey_2023}.

Conventional drift detection relies on statistical monitoring that provides limited insight into the \emph{causal mechanisms} driving degradation \citep{goncalves_comparative_2014, scholkopf2021toward}. Similarly, robustness evaluation via noise injection breaks causal structure, creating counterfactual data points outside the manifold of possible observations \citep{arjovsky2019invariant, chen_estimating_2022}.


To address these limitations, we propose a ``Digital Twin'' framework \citep{mihai_digital_2022} based on interpretable Structural Causal Models (SCM) \citep{pearl_causality_2009} that complements statistical monitoring with preemptive, scenario-based stress-testing. Unlike black-box generative models, it preserves structural dependencies while enabling \textit{Causal Parametric Drift Simulation}, which incrementally shifts specific causal mechanisms to expose latent vulnerabilities that statistical monitors alone cannot anticipate. By grounding robustness evaluation in interventional reasoning and causal discovery, our work demonstrates that causal (not merely statistical) knowledge is essential for anticipating classifier failure modes. Code and experimental scripts are publicly available at \url{https://github.com/Julien-Lafrance/Causal-Parametric-Drift-Simulation}.

The rest of this paper is structured as follows. We first review limitations of standard evaluation and the use of generative models in robustness testing. Section~\ref{s:f-cdt} details our framework, introducing SCM-based Digital Twins and Causal Parametric Drift Simulation. Section~\ref{s:exp_clear} describes the experimental setup, including dataset details, twin construction, validation, and a sanity check on the LUCAS dataset. Section~\ref{s:results} presents the main robustness results on the Open Sourcing Mental Illness (OSMI) dataset, including bootstrap confidence intervals and complementary performance metrics. Section~\ref{s:comparisons} compares the framework against unsupervised and supervised drift monitors, SHAP-based feature importance, replacement-noise baselines, and assesses sensitivity to the discovered causal structure. Finally, we discuss the paradigm shift toward active causal diagnostics and address limitations such as the \textit{Rashomon Set} \citep{semenova_existence_2022} and linearity constraints.
\section{Related Work}

\subsection{Concept Drift and Static Evaluation}
Standard model evaluation relies on static hold-out sets that assume stationarity \citep{ashmore_assuring_2022}. In practice, deployed models encounter \textit{concept drift}: shifts in the joint distribution $P(X,Y)$ where $P(Y|X)$ evolves while $P(X)$ may remain stable \citep{lu_learning_2018, zliobaite2016overview, tsymbal2004problem}. Detecting such changes is difficult \citep{lu_learning_2018}: conventional detectors monitor marginal distributions (e.g., KS tests) and fail to flag parametric shifts in the decision boundary when input statistics remain stable \citep{goncalves_comparative_2014}. Multivariate approaches such as PCA-based reconstruction error \citep{SOUIDEN2022100463} and autoencoders \citep{yong2020bayesian} capture shifts in $P(X)$ but still cannot explicitly track $P(Y|X)$ when ground-truth labels lag, and thus may fail to distinguish benign distributional shifts from genuine concept drift \citep{gemaque2020overview}.

\paragraph{Explainability and Robustness in Tabular Data.}
Post-hoc explainability methods such as LIME \citep{ribeiro_why_2016} and SHAP \citep{lundberg_unified_2017} are fundamentally correlational \citep{janzing2020feature}: their marginal imputations disrupt the joint distribution, producing explanations based on causally impossible counterfactuals \citep{Kumar2020Problems, frye2020asymmetric}. Standard robustness evaluation via replacement noise \citep{chuah_framework_2022} similarly ``breaks the data dependency structure'' \citep{SULLIVAN2021530}, conflating model fragility with confusion over causally invalid inputs (we demonstrate this empirically in Section~\ref{ssb:replacement}). Unlike adversarial robustness (where off-manifold perturbations are acceptable threats), concept drift robustness requires evaluating plausible, causally valid shifts in the data-generating process, preserving the \emph{causal} manifold of the data.

\subsection{Generative Models}
Deep generative models such as GANs \citep{Goodfellow2014Generative} and VAEs \citep{Kingma2013Auto} achieve high fidelity in sampling $P(X,Y)$ \citep{xu2019modeling}, but operate as black boxes with entangled latent spaces that preclude controlled interventions \citep{sehwag_robust_2022}. Causally-aware variants such as CausalGAN \citep{kocaoglu2018causalgan}, Causal-TGAN \citep{wen2021causaltgan}, and Deep SCMs \citep{pawlowski2020deep} incorporate graph structure for counterfactual generation, but their neural components still limit the \textit{parametric transparency} needed for interpretable, mechanism-level interventions.

\paragraph{Digital Twins.}
Digital Twins are simulation environments originally conceptualized for industrial engineering to mirror physical processes \citep{grieves2023digital}. Unlike standard generative models, they support predictive failure analysis by maintaining a dynamic representation of a system's state, relying on explicit parameters to test hypotheses. However, their application has been restricted to deterministic or physically grounded domains such as logistics or autonomous driving \citep{yu2025addt}. Adapting this paradigm for tabular classification, where the environment is abstract and probabilistic, requires replacing physical laws with \textit{causal laws}, yielding a structurally interpretable ``white-box'' generative system.

\paragraph{Causal Discovery and Structural Causal Models.}
Causal inference provides the formalism to distinguish observational $P(Y|X)$ from interventional $P(Y|do(X))$ \citep{pearl_causality_2009}. Causal Discovery algorithms such as the Peter-Clark (PC) algorithm \citep{spirtes_causation_2001} recover causal graphs from observational data via conditional independence testing under the Causal Markov Condition and \textit{faithfulness} \citep{glymour_review_2019}. Alternative approaches jointly recover structure and parameters: LiNGAM \citep{shimizu_linear_2006} exploits non-Gaussianity, while NOTEARS \citep{zheng2018notears} formulates discovery as continuous optimization. Our framework is agnostic to the specific discovery method; the two-step pipeline reflects an implementation choice for modularity.

However, discovery algorithms optimize for topological consistency rather than parametric fit. Once a graph is identified, Structural Causal Models (SCM) offer a framework to estimate parameters for the mixed-type and categorical nature of real-world tabular data \citep{pearl2009causal}. On the other hand, Structural Equation Models literature, which is an inspiration for SCMs, focuses on \textit{global} goodness-of-fit given a structure known a priori. They do so by evaluating how well a model reproduces the expected population covariance matrix $\Sigma$ with their own covariance matrix $S$ given the sample size $N$ and the number of observed variables $p$. \citet{schermelleh2003evaluating} highlight metrics such as $\chi^2$ and the Root Mean Square Error of Approximation ($RMSEA$) based on $\chi^2$ and the degrees of freedom $df$:

\begin{equation}
    \chi^2 = (N - 1) \left[ \ln|\Sigma| - \ln|S| + \text{tr}(S\Sigma^{-1}) - p \right]
\end{equation}
\begin{equation}
    RMSEA = \sqrt{\max\left(0,\frac{ \chi^2 - df}{df(N - 1)}\right)}
    \label{rmsea}
\end{equation}

We note that the integration of causally discovered models and SEM evaluation has precedents in specific domains. For instance, Stable Specification Search \citep{rahmadi2017causality} and packages like \texttt{SEMgraph} \citep{SEMgraph} utilize fit indices to refine causal graphs in biological and psychological research. However, we argue that these rigorous validation loops are an important tool to operationalize faithful Digital Twins for machine learning. Unlike physical domains where twins are validated against immutable laws of physics, tabular data requires a validation framework that can simultaneously discover and verify the \textit{causal laws} of the environment. Consequently, bridging the gap between causal structure learning through causal discovery and generative validation through SEM goodness-of-fit metrics is the critical step required to move from black-box generative models to trustworthy interpretable hypothesis-enabling Digital Twins.
\section{Framework: Causal Digital Twins for Robustness Evaluation}
\label{s:f-cdt}
To address the limitations outlined above in robustness evaluation and ad-hoc explainability, we propose a framework that operationalizes the concept of a Digital Twin (DT) for tabular classification pipelines. We use ``Digital Twin'' in a specific, limited sense: a simulation environment with known limitations, not a perfect replica of reality. The objective is to obtain a useful proxy of the environment that enables hypothesis-driven stress testing.

Our framework inherits standard assumptions from causal discovery and SEM: (i)~\textit{causal sufficiency}, i.e., no unmeasured confounders (relaxable via the FCI algorithm \citep{spirtes_causation_2001}, at the cost of less specific output); (ii)~\textit{faithfulness}; (iii)~the \textit{Causal Markov Condition}; (iv)~\textit{acyclicity}; and (v)~\textit{linearity} of the structural equations $f_v$ (see Phase~2 below). Violations of (i)--(iv) may produce incorrect edges or biased estimates; violation of~(v) limits generative fidelity. However, empirical validation (Phase~3) and our sensitivity analysis (Section~\ref{ssb:sensitivity}) show that Breaking Points remain stable across different graph structures, providing partial robustness to moderate violations.

\begin{figure}[H]

\centering

\definecolor{warranted}{HTML}{0FBF63}
\definecolor{actual}{HTML}{FF914D}
\definecolor{perceived}{HTML}{004AAD}
\begin{tikzpicture}[>=latex,scale=1]
\node[draw] at (-0.5,0) (1) {Observed Dataset};
\node[draw] at (8,0) (A) {Domain Constraints};

\node[draw, below=0.335cm  of 1.south] (1-1) {Classification\text{\color{white}p}Pipeline};
\draw[->] (1.south) -- (1-1.north);
\node[draw, right=1.6cm of 1] (1-3) {Causal Graph};
\draw[->] (1) -- (1-3);
\draw[->] (A) -- (1-3);

\node[above left=-0.08cm and -0.9cm of 1-3] (DT) {};
\node[draw, below=0.3cm of 1-3.south] (2-1) {Structural\text{\color{white}p}Causal\text{\color{white}p}Model};
\draw[->] (1-3) -- (2-1);
\draw[->] (1) -- (2-1);

\node[draw, below=0.4cm of 2-1.south] (3-1) {Generated Test Datasets};
\draw[->] (2-1) -- (3-1);
\node[draw, right=0.6cm of 3-1.east] (3-2) {Scenario Definition};
\draw[->] (3-2) -- (3-1);
\node[draw, below=0.4cm of 1-1.south] (4-1) {Evaluations};
\draw[->] (1-1.south) -- (4-1);
\draw[->] (3-1) -- (4-1);

\end{tikzpicture}
\caption{Structure of the Digital Twin framework and its inputs.}
\label{f:structure_dt}
\end{figure}
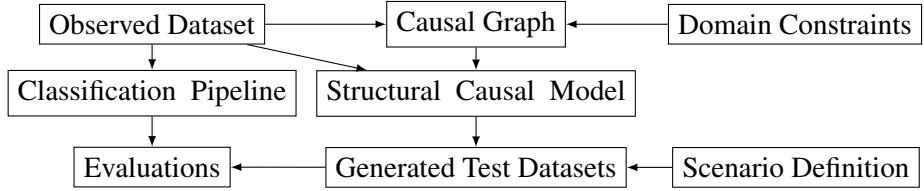

In this context, a SCM  acts as the generative engine of the DT that enables explicit parameter manipulation while replicating as faithfully as possible the causal mechanisms and statistical dependencies of the observational data. Figure \ref{f:structure_dt} presents the general structure of the DT, its inputs and interactions between each of its components.

\subsection{Operationalizing the Digital Twin with SCMs}
\label{sb:operationalization}
The core of our framework is an SCM discovered from observational data. We leverage the capability of modern SCMs to handle discrete and categorical variables not merely for estimation of causal impacts, but for controlled data generation. Formally, we represent the data-generating process as a tuple $\mathcal{M} = \langle V, U, \mathcal{F}, P(U) \rangle$, where:
\begin{itemize}
    \item $V = \{v_1, \dots, v_p\}$ is the set of observed variables including features and the target variable.
    \item $U = \{u_1, \dots, u_p\}$ is the set of noise terms.
    \item $\mathcal{F} = \{f_v\}_{v \in V}$ is a set of structural equations, where each $v_i \leftarrow f_i(Pa_i, u_i)$ is determined by its causal parents $Pa_i \subset V$ and an error term $u_i$.
    \item $P(U)$ is the joint probability distribution over the exogenous variables.
\end{itemize}

To ensure the Digital Twin serves as a reliable proxy, we employ a hybrid discovery-validation pipeline that systematically instantiates the components of $\mathcal{M}$:

\paragraph{Phase 1: Structure Learning (Defining $V$ and Arguments of $\mathcal{F}$)}
We employ a causal discovery algorithm to recover the causal structure of the observed dataset as a Directed Acyclic Graph. Specific implementations of this algorithm may vary, ranging from constraint-based methods that utilize conditional independence tests \citep{spirtes_causation_2001} to score-based approaches. 
The chosen algorithm must also support the injection of domain-specific background knowledge like restricting nodes to being exogenous or forbidding certain edges. This approach restricts the search space, significantly raising computational efficiency while ensuring the resulting topology respects known physical or logical dependencies that statistical metrics alone might miss \citep{zhang2023enhancing}. Note that constraint-based discovery may return a Completed Partially Directed Acyclic Graph (CPDAG) rather than a fully oriented DAG; the domain constraints described above typically resolve remaining edge orientations, as was the case in our experiments.

Crucially, this step identifies the observed variable set $V$ directly from the data. We define the required structural arguments for the function set $\mathcal{F}$, ensuring the topology respects local conditional independence constraints. The resulting Directed Acyclic Graph serves as the topological blueprint for the Digital Twin Environment, defining which variables are causal parents of others ($Pa_i \subset V$).

\paragraph{Phase 2: Structural Equation Fitting (Defining Mechanisms of $\mathcal{F}$, $U$, and $P(U)$).} 
Once the topology is fixed, we must define the functional form of the relationships $f_v$. Standard SEM implementations often default to linear assumptions, which are ill-suited when tabular data is of categorical nature, thus justifying our use of SCMs. To maintain both \textit{interpretability} and \textit{generative validity}, we explicitly define three possible forms that $f_v$ can take (note: these linearity constraints apply only to the SCM's structural equations, not to the classifier under test, which can be arbitrarily complex):
\begin{itemize}
    \item \textbf{Endogenous Numeric Variables:} For continuous features with $|Pa_i| > 0$, we model dependencies via linear combinations of its parents with $u_i = \mathcal{N}(0, \sigma^2)$.
    \item \textbf{Endogenous Categorical Variables} For discrete features with $|Pa_i| > 0$ and $K$ possible output values of $f_v$, we employ a generalized logistic framework to ensure valid probabilistic outputs.
    By defining all endogenous variables with $f_v$ as regressions, we are able to represent the mechanism's parameters (Table \ref{f:output_permutation_viz}) to facilitate the modification and interpretation of $\mathcal{M}$ :

    \begin{itemize}
        \item \textbf{Binary Case ($K=2$):} We utilize standard Logistic Regression with a sigmoid link function with Gaussian noise introduced before the sigmoid link to estimate $P(k=2)$. If $P(k=2) > 0.5$, then $k=2$, inversely, if $P(k=2) < 0.5$, then $k=1$.
        \item \textbf{Multinomial Case:} We extend this to Multinomial Logistic Regression using the softmax function. The model estimates $K$ parameter sets to produce a normalized probability vector $[p_1, ..., p_K]$ such that $\sum p_k = 1$. The output of $f_v$ is the category $k$ with the highest $p_k$. The Gaussian noise is sampled $K$ times and applied to each linear combination before the application of the Softmax function \citep{softmax_jingli}. 
        \item \textbf{Exogenous Root Nodes:} For variables with no parents ($Pa_i = \emptyset$), the ``structural equation'' is simply the generation of the background variable itself. We model $f_{u\in U}$  using Gaussian Mixture Models for numeric data or empirical sampling for categorical data.
        This formulation is critical for the Digital Twin's operation: explicitly modelling the probability distribution of root nodes allows us to modify the generation mechanisms by modifying the mean and variance of the Gaussians and their weights in the mixture, or by modifying the frequencies in the empirical distribution.
    \end{itemize}
\end{itemize}
    \begin{table}[h]
        \centering
        \setlength{\fboxsep}{2.5pt}

        \begin{tabular}{@{}c|ccc@{}}
             \textbf{Outputs}& \textbf{$Pa_0$} & \textbf{$Pa_1$}&\textbf{Intercept}\\
             \midrule
             \textbf{$y_0$} & $\beta_{0,0}$ & $\beta_{0,1}$ & $b_0$\\
             \textbf{$y_1$} & $\beta_{1,0}$ & $\beta_{1,1}$ & $b_1$\\
             \textbf{$y_2$} & $\beta_{2,0}$ & $\beta_{2,1}$  & $b_2$\\
        \end{tabular}
        $\xrightarrow{}$
        $\begin{bmatrix} 
            \beta_{0,0} & \beta_{0,1} &b_{0}\\
            \beta_{1,0} & \beta_{1,1}&b_{1}\\
            \beta_{2,0} & \beta_{2,1}&b_{2}\\
        \end{bmatrix}$
        \caption{Visualization of the parameters of the data generation function $f_v$ of a multiclass node with two parent nodes.}
    \label{f:output_permutation_viz}
    \end{table}
This phase completes the definition of $\mathcal{M}$ by populating $U$ with valid noise terms, defining $P(U)$ for stochastic generation, and fitting all data generation mechanisms $\mathcal{F}$. Crucially, each $f_v$ yields interpretable parameters that practitioners can modify to shift the data structure of generated observations.

\paragraph{Phase 3: Global Generative Validation}
This step ensures that the synthetic datasets $\mathcal{D}_{gen}$ are as statistically indistinguishable as possible from the real distribution $\mathcal{D}_{real}$ before any intervention is applied. Since the Causal Markov Condition is assumed to be satisfied during Structure Learning, our validation focuses on the fidelity of the generated samples rather than topological independence testing. We employ a three-tiered validation protocol to accept the estimated model $\mathcal{M}$ as a valid Digital Twin Environment:

\begin{enumerate}
    \item \textbf{Global Structural Fit (Covariance Alignment):} We first verify that the model's covariance matrix $\Sigma$ calculated from $\mathcal{M}$ approximates the observed covariance matrix $S$ by calculating the Root Mean Square Error of Approximation (Equation \ref{rmsea}).

    We accept the Digital Twin only if $RMSEA \le 0.08$. We note that this threshold is slightly more lenient than the strict 0.06 often cited in psychometrics \citep{ullman2012structural}. This relaxation is intentional. We base this threshold choice on the recommendation by \citet{Hooper2008StructuralEM} to have the maximum upper bound of $RMSEA$ to be lower than 0.08. This choice is justified since our objective is to optimize for causal representativeness in tabular datasets and to evaluate robustness rather than a theoretical fit to explain the phenomenon.

    \item \textbf{Marginal Distributional Fidelity:} A low $RMSEA$ ensures global structural alignment but does not guarantee that individual feature distributions are preserved, particularly for multimodal or skewed variables. We validate the univariate distributions by comparing $\mathcal{D}_{gen}$ and $\mathcal{D}_{real}$ using the Kolmogorov-Smirnov test \citep{kolmo_smirnov} for continuous variables and Cramer's V \citep{chi-square-cramer-v} for categorical variables. We do not directly use the $\chi^2$ test for categorical features, since the sample size necessary for fitting our model will generally accept the null hypothesis. We require that the generated marginals do not statistically diverge from the observations ($p > 0.05$) or $V < 0.1$ \citep{cohen1988statistical}, ensuring the Digital Twin Environment initially generates values strictly within the valid domain of the data manifold.

    \item \textbf{Predictive Consistency:} Finally, we assess the preservation of the conditional distribution $P(Y|X)$ by evaluating the utility of the data for downstream tasks. We train a classifier on $\mathcal{D}_{real}$ and evaluate its performance on $\mathcal{D}_{valid}$ and $\mathcal{D}_{gen}$. A Digital Twin Environment should yield performance metrics comparable to those obtained on the real hold-out validation set, providing additional confirmation that the decision boundary landscape has been faithfully replicated. If any tier fails, practitioners should refine the structural constraints or $f_v$ specification before proceeding.
\end{enumerate}

\subsubsection{Causal Parametric Drift Simulation Through Hypothesis Testing}
\label{ssb:cpds}
Real-world concept drift is rarely random; it often involves the strengthening or weakening of specific relationships. Our framework allows practitioners to test specific relationship hypotheses by applying relative shifts to the parameters of $f_v$ from $\mathcal{M}$:
$$
\beta'_{ij} = (1 + \delta_{ij})\,\beta_{ij}
$$
where $\delta_{ij}$ is the relative drift magnitude (e.g., $\delta = -0.35$ represents a 35\% weakening). By causally generating synthetic test datasets across a range of $\delta_{ij}$, we can profile the \textbf{Robustness Curve} of the classifier. This reveals the \textbf{Breaking Point} of the model, the exact magnitude of causal shift required to degrade performance below an acceptable threshold given a specific drift scenario. A hypothesis test can contain $\delta_{ij}$ for multiple nodes, which makes it possible to test complex hypotheses.

This approach shifts the evaluation paradigm from passive observation of error metrics to active, scenario-based diagnostics, allowing for the anticipation of failure modes before they occur in deployment. To produce a  \textbf{Robustness Curve} mapping the drift magnitude to model performance, we execute the following procedure:
\begin{enumerate}
    \item \textbf{Baseline Generation:} Generate a baseline synthetic dataset $\mathcal{D}_{0}$ using the causally discovered SCM with its original parameters $\beta$ for each node.
    \item \textbf{Incremental Intervention:} For a specific target relationship or set of relationships, incrementally vary $\delta_{ij}$ through the range of values to test (e.g. $\delta_{ij}$ from $-0.5$ to $1.0$).
    \item \textbf{Twin Generation:} At each time step $k$, apply to SCM parameters their respective $\delta_{ij}$ and sample a drifted dataset $\mathcal{D}_{k}$.
    \item \textbf{Evaluation:} Evaluate the classifier $\mathcal{C}$ on $\mathcal{D}_{k}$ to obtain a performance metric $M_k$.
\end{enumerate}

\paragraph{The Breaking Point}
The critical output of this analysis is the \textbf{Breaking Point}, defined as the minimum relative drift $\delta_{crit}$ required to push the classifier $\mathcal{C}$'s performance metric $M$ below a safety threshold $\tau$ chosen given the application domain:
\begin{equation}
    \delta_{crit} = \min_{\delta} \{ |\delta| : M(\mathcal{C}, \mathcal{D}_{k}) < \tau \}
\end{equation}
For instance, $\delta_{crit} = -0.35$ indicates the model tolerates up to a 35\% weakening of the targeted mechanism before crossing the safety threshold. This shifts evaluation from passive monitoring to active, anticipatory diagnostics with concrete thresholds to monitor.

\section{Experimental Setup}
\label{s:exp_clear}
To validate the efficacy of our Causal Digital Twin framework, we conducted a series of experiments using real-world tabular datasets in \texttt{Python}. The objective is not to maximize predictive performance, but to use the Digital Twin to expose latent vulnerabilities of a standard ``black-box'' classifier under controlled causal interventions. We organize the experiments around two datasets: a synthetic-truth benchmark (LUCAS) used to verify that the discovery and fitting pipeline behaves as expected when the causal structure is known, and the Open Sourcing Mental Illness (OSMI) Mental Health in Tech survey dataset, on which we run the main robustness study.

\subsection{Datasets and Preprocessing}
\label{ssb:datasets}

\paragraph{Open Sourcing Mental Illness (OSMI).}
The OSMI dataset is the Open Sourcing Mental Illness \emph{Mental Health in Tech} survey, which contains data on mental-health conditions in the technology workplace, with 23 features covering demographics, employer policies, attitudes toward mental health, and self-reported outcomes. We selected it for three reasons: (i)~it has interpretable latent causal structure suitable for SEM-style modelling; (ii)~it is dominated by categorical variables, requiring SCMs that handle non-continuous distributions; and (iii)~as noted by \citet{m_s_survey_2023}, healthcare is a domain where trust and robustness are critical. The classification target is the binary feature \texttt{treatment} (whether the respondent has sought treatment for a mental-health condition). Categorical variables were encoded numerically for the classifier and SCM estimation, \texttt{Age} was min--max normalized to $[0,1]$, and the only feature with missing values, \texttt{work\_interfere}, had its missing class promoted to its own category since the survey phrasing assigns it a semantically distinct meaning. The full data dictionary, including each feature's possible values and the corresponding survey question, is provided in Appendix~\ref{a:osmh-dataset} (Tables~\ref{tab:desc-dataset-1}--\ref{tab:desc-dataset-2}).

\paragraph{LUCAS.}
LUCAS \citep{lucas_dataset} is a small synthetic dataset for which the ground-truth causal topology is known. We use it as a sanity check to confirm that our implementation of the PC algorithm and the generalized logistic regression framework recover the correct structure and produce a faithful generative model on a less complex problem. The full LUCAS validation appears in Section~\ref{ssb:lucas}.

\subsection{Target Models and Digital Twin Construction}
\label{ssb:setup-models}
\paragraph{Target Models.}
We trained a gradient-boosted decision tree predictor for the \texttt{treatment} variable. The \texttt{XGBoost} package \citep{xgboost_2016} was chosen for this classifier due to its prevalence in tabular competitions and industrial applications, and its opacity relative to linear models. We also trained a Random Forest model from \texttt{scikit-learn} \citep{scikit-learn} as a comparison. Both models used default hyperparameters and were evaluated with cross-validation on an 80--20 split of the target variable $Y$ \texttt{treatment}, achieving baseline F1 scores of $0.797$ and $0.811$ on $\mathcal{D}_{valid}$.

\begin{figure}[H]
    \centering
    \includegraphics[width=1\linewidth]{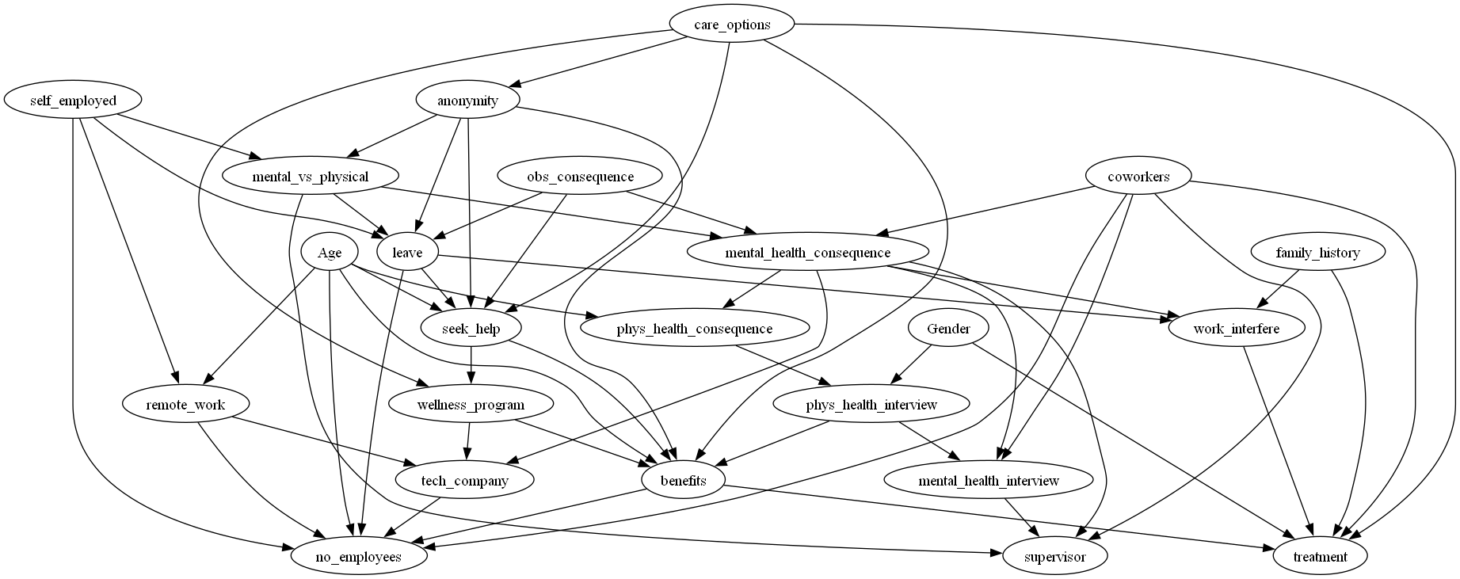}
    \caption{Causal graph topology discovered by the PC algorithm on the OSMI dataset with $\alpha_{PC}=0.05$ and our domain constraints.}
    \label{fig:causal_graph}
\end{figure}

\paragraph{Digital Twin Construction.}
We constructed the SCM for the Digital Twin in two phases using the \texttt{DoWhy} Python library \citep{sharma2020dowhyendtoendlibrarycausal, dowhy_blaum}:
\begin{enumerate}
    \item \textbf{Structure Learning:} We employed the Peter--Clark algorithm \citep{spirtes_causation_2001} with a significance threshold of $\alpha=0.05$ to discover the causal graph from the observational dataset. We applied domain constraints to enforce \texttt{Age}, \texttt{Gender}, \texttt{self\_employed} and \texttt{family\_history} as exogenous source nodes and \texttt{treatment} as a sink node. The discovered graph appears in Figure~\ref{fig:causal_graph}; sensitivity to these choices is examined in Section~\ref{ssb:sensitivity}.
    \item \textbf{Parameter Estimation:} We fit the functional relationships of each node according to the criteria defined in Section~\ref{sb:operationalization}. Continuous variables were modelled with linear regressions; categorical variables were modelled with the generalized logistic regression framework to ensure probabilistic validity. These regressions were implemented using \texttt{scikit-learn} \citep{scikit-learn} with default hyperparameters.
\end{enumerate}

\subsection{Generative Model Validation}
\label{ssb:gen-validation}
To ensure the reliability of our simulations, we validated the unmodified Digital Twin ($\mathcal{M}$) using the three-tiered protocol defined in Section~\ref{sb:operationalization}.

\begin{figure}[H]
    \centering
    \includegraphics[width=0.8\linewidth]{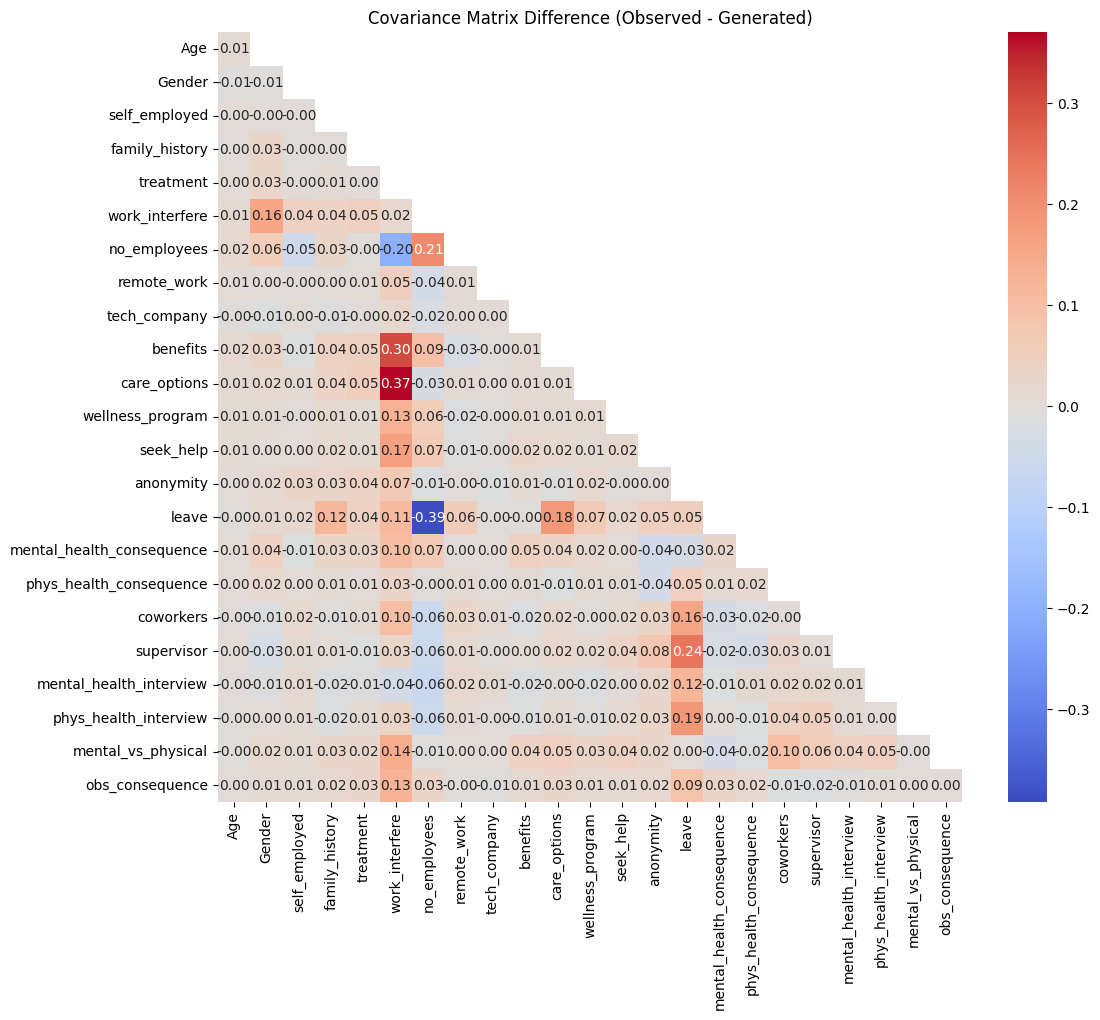}
    \caption{Difference between the observed covariance matrix of the OSMI dataset and the covariance matrix of data generated by the unmodified Digital Twin.}
    \label{fig:diff_cov_osmh}
\end{figure}

\paragraph{Global Structural Fit.}
The fitted SCM achieved an RMSEA of $0.0678$, slightly under our tolerance threshold of $0.08$, indicating an adequate correspondence between the implied and observed covariance structures. To calculate the number of degrees of freedom, we employed the strategy from \citet{nielsen_calculating_2025}, which gave $df=182$. Figure~\ref{fig:diff_cov_osmh} shows the elementwise difference between the observed and generated covariance matrices: most entries fall close to zero, consistent with the favourable RMSEA value.

\paragraph{Marginal Fidelity.}
We verified univariate distributions using Kolmogorov--Smirnov tests for continuous variables and Cramer's V for categorical variables. All features satisfied the consistency requirement ($p > 0.05$), confirming that the generative process preserves the data manifold without introducing spurious drift.

\paragraph{Predictive Consistency.}
As shown in Table~\ref{tab:predictive_consistency}, we assessed downstream utility by evaluating both classifiers on the validation dataset $\mathcal{D}_{valid}$ and on data $\mathcal{D}_{gen}$ generated by the unmodified twin. The performance gap is negligible, confirming that the Digital Twin faithfully replicates the decision-boundary landscape of the original environment.

\begin{table}[H]
\centering
\begin{tabular}{lcccc}
\toprule
\textbf{Models} & \textbf{Accuracy} & \textbf{Precision} & \textbf{Recall} & \textbf{F1} \\
\midrule
XGBoost ($\mathcal{D}_{valid}$) & $0.797 \pm 0.012$ & $0.796 \pm 0.018$ & $0.798 \pm 0.015$ & $0.797 \pm 0.014$ \\
XGBoost ($\mathcal{D}_{gen}$) & $0.794 \pm 0.004$ & $0.796 \pm 0.004$ & $0.793 \pm 0.004$ & $0.794 \pm 0.004$ \\
\midrule
Random Forest ($\mathcal{D}_{valid}$) & $0.811 \pm 0.008$ & $0.809 \pm 0.008$ & $0.812 \pm 0.008$ & $0.810 \pm 0.008$ \\
Random Forest ($\mathcal{D}_{gen}$) & $0.802 \pm 0.003$ & $0.806 \pm 0.003$ & $0.801 \pm 0.002$ & $0.803 \pm 0.003$ \\
\bottomrule
\end{tabular}
\caption{Performance metrics on OSMI validation data and on unmodified-SCM-generated data. The minimal drop between $\mathcal{D}_{valid}$ and $\mathcal{D}_{gen}$ falls inside the standard-deviation bands, corroborating the Digital Twin's predictive consistency.}
\label{tab:predictive_consistency}
\end{table}

\subsection{Validation on a Known Topology: LUCAS}
\label{ssb:lucas}
\begin{figure}[H]
    \centering
    \begin{minipage}[b]{0.48\linewidth}
        \centering
        \includegraphics[width=\linewidth]{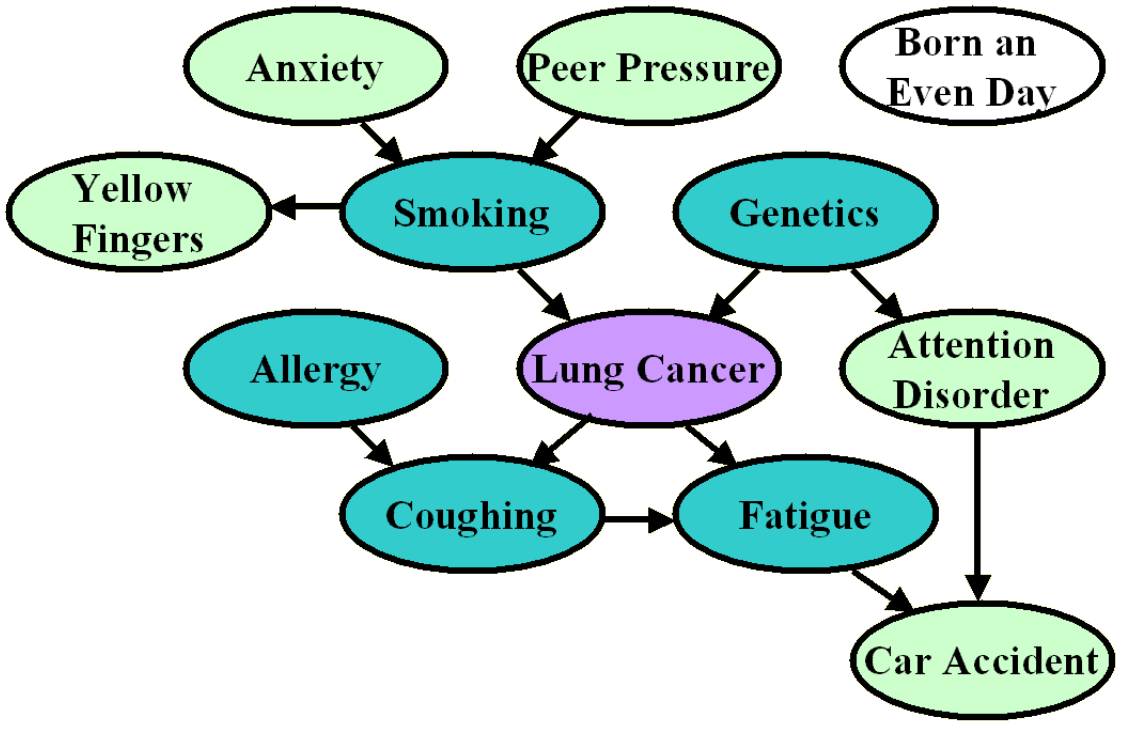}
        \\[0.3em](a) Published ground truth.
    \end{minipage}\hfill
    \begin{minipage}[b]{0.48\linewidth}
        \centering
        \includegraphics[width=\linewidth]{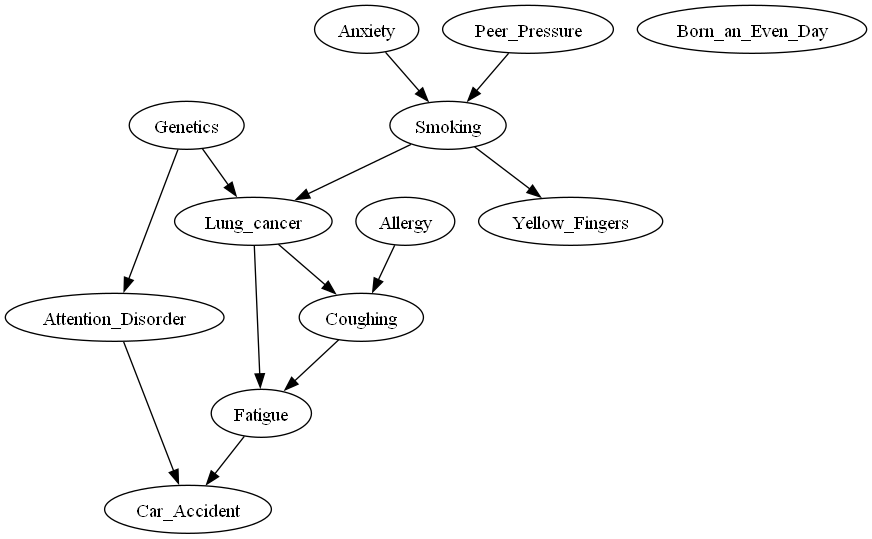}
        \\[0.3em](b) Recovered by PC at $\alpha=0.05$.
    \end{minipage}
    \caption{LUCAS causal graph: published ground truth (a, \citep{lucas_dataset}) versus the topology our pipeline discovers from the data alone (b). The two are identical up to node placement.}
    \label{fig:lucas_comparison}
\end{figure}

Because OSMI does not have a ground-truth causal graph against which we can compare our discovered structure, we ran the same construction-and-validation pipeline on the LUCAS dataset \citep{lucas_dataset}, which has a known topology. Figure~\ref{fig:lucas_comparison} places the published ground truth (a) next to the topology our pipeline discovers from the data alone with the PC algorithm at $\alpha=0.05$ (b). The two graphs are identical up to node placement, confirming that the discovery step recovers the correct structure when one is known.

\begin{figure}[H]
    \centering
    \includegraphics[width=0.90\linewidth]{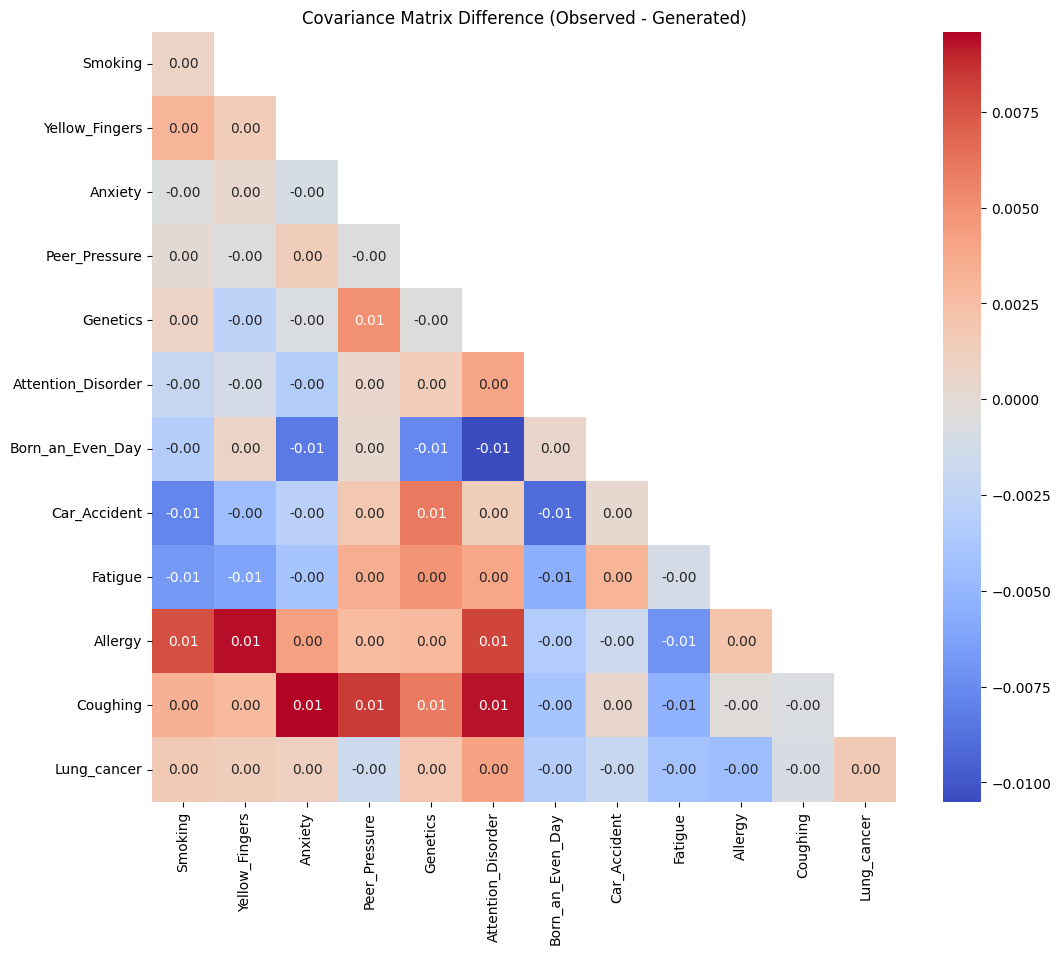}
    \caption{Difference between the observed covariance matrix of the LUCAS dataset and the covariance matrix of data generated by the LUCAS Digital Twin.}
    \label{fig:diff_cov_LUCAS}
\end{figure}

Re-running the three-tiered protocol on this twin gave equally favourable results: an RMSEA of $0.0152$ (well under $0.08$) with $df=44$, all Cramer's V tests above the consistency threshold, and predictive consistency between $\mathcal{D}_{valid}$ and $\mathcal{D}_{gen}$ comparable to OSMI (Table~\ref{tab:predictive_consistency_rf_lucas}). Figure~\ref{fig:diff_cov_LUCAS} shows the corresponding covariance-matrix difference. Together, these results confirm that the discovery and fitting pipeline behaves as expected when the underlying structure is known.

\begin{table}[H]
    \centering
\begin{tabular}{lcccc}
\toprule
\textbf{Models} & \textbf{Accuracy} & \textbf{Precision} & \textbf{Recall} & \textbf{F1} \\
\midrule
XGBoost ($\mathcal{D}_{valid}$) & $0.850 \pm 0.013$ & $0.825 \pm 0.027$ & $0.800 \pm 0.017$ & $0.810 \pm 0.018$ \\
XGBoost ($\mathcal{D}_{gen}$) & $0.849 \pm 0.003$ & $0.821 \pm 0.004$ & $0.797 \pm 0.003$ & $0.807 \pm 0.004$ \\
\midrule
Random Forest ($\mathcal{D}_{valid}$) & $0.842 \pm 0.012$ & $0.809 \pm 0.024$ & $0.791 \pm 0.005$ & $0.798 \pm 0.008$ \\
Random Forest ($\mathcal{D}_{gen}$) & $0.837 \pm 0.003$ & $0.807 \pm 0.004$ & $0.788 \pm 0.002$ & $0.797 \pm 0.004$ \\
\bottomrule
\end{tabular}
\caption{Performance metrics on LUCAS validation data and on unmodified-SCM-generated data. The minimal gap corroborates the Digital Twin's predictive consistency on a dataset with known ground-truth structure.}
\label{tab:predictive_consistency_rf_lucas}
\end{table}

\section{Results: Robustness Curves and the Breaking Point}
\label{s:results}
With the Digital Twin validated, we now apply Causal Parametric Drift Simulation (Section~\ref{ssb:cpds}) to identify the \textbf{Breaking Point} of the XGBoost classifier under a plausible causal shift in the OSMI environment. A secondary objective is to show that statistical monitoring fails when the \emph{semantic} sense of a feature shifts while its \emph{distribution} does not.

\subsection{Scenario Definition: ``Self Help''}
\label{ssb:scenario}
We hypothesized a ``Self Help'' scenario in which employees increasingly perceive professional mental-health treatment as inaccessible or non-beneficial. Consequently, while they continue to report that their mental health interferes with their work, they become less likely to seek medical help, opting instead for self-treatment \citep{oracle2020ai}.\footnote{Cross-sectional data cannot establish temporal precedence \citep{hernan2020causal}. However, the survey's phrasing supports treating interference as antecedent to treatment-seeking. Even if the true direction were uncertain, identifying this relationship as a key vulnerability remains actionable.}
In the SCM, this manifests as a weakening of the positive causal link between \texttt{work\_interfere} and \texttt{treatment}. Since their covariance is $0.86$, we anticipate that the classifier has learned a strong link between work interference and treatment, and will suffer degradation as this link weakens, generating false positives. We establish the safety threshold $\tau_{Precision}$\footnote{We prefer Precision over F1, which has known limitations as a single-summary evaluation metric in deployment settings, notably its insensitivity to true negatives \citep{vancalster2025evaluation}; complementary metrics are reported in Section~\ref{ssb:multiple-metrics}.} based on the minimum viable utility for a screening tool. Given the prevalence of treatment-seeking in the dataset $(P(Y=1) \approx 0.5)$, a random guess yields a precision of $0.5$. To illustrate the framework, we require a safety margin of 40\% over this random-guess baseline, yielding $\tau_{Precision}=0.5\times1.4=0.7$; the qualitative conclusions of this paper are insensitive to small variations in this threshold (Section~\ref{ssb:multiple-metrics}). Falling below $\tau_{Precision}$ indicates the model is not significantly more reliable than a coin flip for positive predictions, rendering it hazardous for high-stakes HR allocation.

\subsection{Procedure}
\label{ssb:procedure}
\begin{figure}[H]
    \centering
    \includegraphics[width=1.00\linewidth]{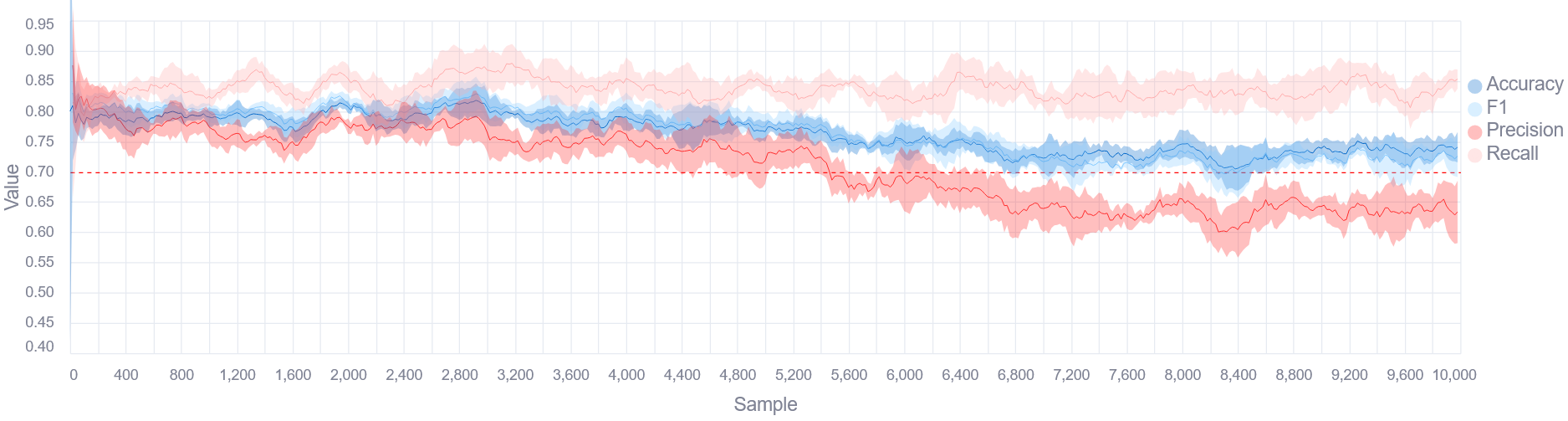}
    \vspace{-0.65cm}
    \caption{XGBoost classifier \textbf{Robustness Curve} in the Self Help scenario. Metrics are computed as rolling-window averages (window size $=300$). $\mathcal{D}_0$ for the first 3000 samples, then $\mathcal{D}_1$ to $\mathcal{D}_{k-1}$ for 4000 samples, and finally $\mathcal{D}_K$ for 3000 samples. The red dashed line marks $\tau_{Precision} = 0.7$.}
 \label{fig:xgboost_robustness_curve}
\end{figure}

We operationalized this hypothesis by targeting the $\delta_{ij}$ parameters (Section~\ref{ssb:cpds}) in the \texttt{treatment} node corresponding to the \texttt{work\_interfere} input, setting $\delta_{\max}=-0.5$. With $K$ drift steps, the step-specific drift $\delta_k$ is:
\begin{align}
        \beta_k = (1 + \delta_k)\,\beta_0  & &\delta_k = \frac{k\,\delta_{\max}}{K}
    \label{delta_k}
\end{align}
We generated a robustness curve by varying $\delta_k$ from $0$ to $\delta_{\max}$ over $K = 20$ steps, following the protocol defined in Section~\ref{sb:operationalization}:
\begin{enumerate}
    \item \textbf{Baseline Generation:} A control dataset $\mathcal{D}_{0}$ was generated from the unmodified SCM (the first 3000 observations of the curve).
    \item \textbf{Incremental Intervention:} We iterated through 20 time steps, applying the incremental drift in Equation~\ref{delta_k} to the targeted coefficients. For each step $k$, $\mathcal{D}_k$ of 200 observations was sampled, totalling 4000 observations. The final state $\beta_K$ is sampled for 3000 observations.
    \item \textbf{Evaluation:} We recorded a rolling-window average (RWA) of size 300 of the classifiers' performance metrics. The RWA evolution is shown in Figure~\ref{fig:xgboost_robustness_curve}.
\end{enumerate}

\vspace{-0.55cm}

\subsection{Robustness Curve Analysis}
\label{ssb:rc-analysis}
The Robustness Curve in Figure~\ref{fig:xgboost_robustness_curve} reveals a distinct decoupling of performance metrics: while Recall remains stable throughout, Precision drops by approximately $0.15$ as the causal link weakens.

\paragraph{Breaking Point ($\delta_{crit}$).}
For $\delta \in [0, -0.30]$, the model maintains performance above the safety threshold. We identify the breaking point at $\delta_{crit} \approx -0.35$, the point at which Precision consistently falls below $\tau$.

\begin{figure}[H]
    \centering
    \includegraphics[width=0.95\linewidth]{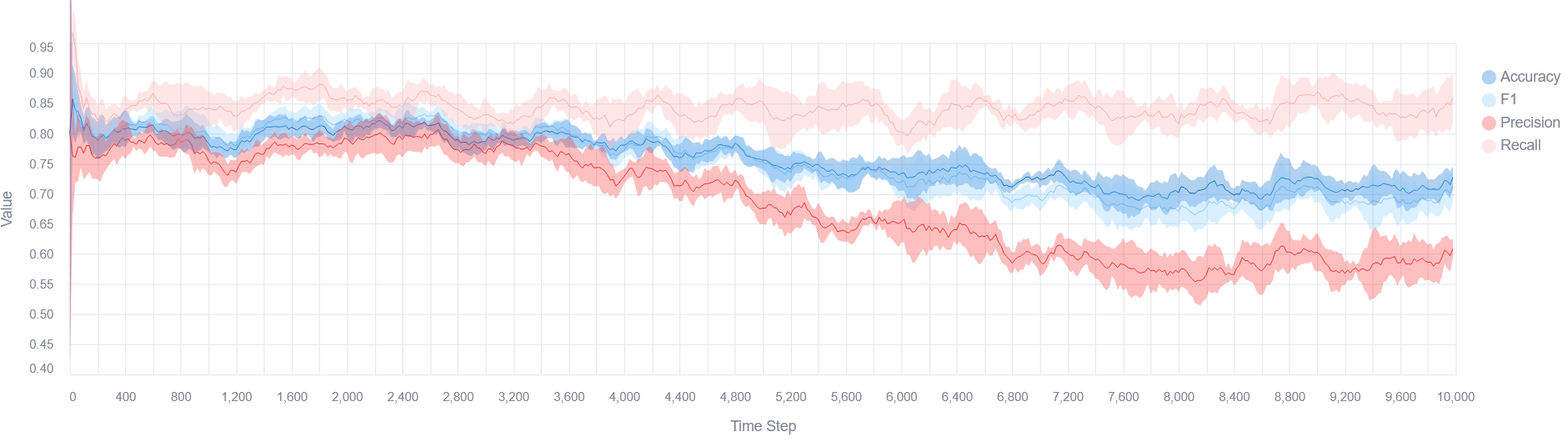}
    \caption{Random Forest classifier Robustness Curve in the Self Help scenario. $\mathcal{D}_0$ for the first 3000 observations, $\mathcal{D}_1$ to $\mathcal{D}_{k-1}$ for 4000 observations, then $\mathcal{D}_K$ for 3000 observations.}
    \label{fig:robust_curve_rf}
\end{figure}

\paragraph{Random Forest comparison.}
The Random Forest classifier exhibits the same qualitative behaviour: Figure~\ref{fig:robust_curve_rf} shows the Self Help scenario applied to the Random Forest, with the same Precision/Recall decoupling and a comparable Breaking Point.

\paragraph{Takeaways.}
This experiment exposes a latent vulnerability: the XGBoost model over-relies on \texttt{work\_interfere} to predict \texttt{treatment}. By identifying $\delta_{crit}$, we quantify the safety margin of the deployed model and demonstrate that a plausible real-world behavioural shift is sufficient to render the model unreliable. As we will show in Section~\ref{ssb:unsupervised}, no unsupervised drift detector raises an alert during this scenario. This is expected: no input distribution shifted, only the conditional $P(\texttt{treatment}\mid\texttt{work\_interfere})$. In this scenario, the diagnostic should push practitioners to track the specific acquisition parameters of \texttt{work\_interfere} so its semantic meaning does not silently change in production, and to reconsider feature selection so the classifier does not use \texttt{work\_interfere} as a proxy for \texttt{treatment}.

\subsection{Bootstrap Confidence Interval for $\delta_{crit}$}
\label{ssb:bootstrap-ci}
Because the GCM's sampling process introduces stochastic variation, the estimated breaking point $\delta_{crit}$ fluctuates across runs. To quantify this uncertainty we conducted $B=50$ bootstrap replications of the Self Help drift simulation, each with a different random seed. For each replication, $\delta_{crit}$ was identified as the first $\delta$ at which Precision fell below $\tau_{Precision}=0.7$ for three consecutive drift steps. Table~\ref{tab:bootstrap_ci} summarizes the resulting distribution: a mean $\delta_{crit}$ near our point estimate, with a 95\% interval that justifies treating the Breaking Point as an order-of-magnitude estimate rather than a precise threshold.

\begin{table}[H]
\centering
\small
\begin{tabular}{rr}
\toprule
Statistic & Value \\
\midrule
Mean $\delta_{crit}$ & $-0.327$ \\
Median $\delta_{crit}$ & $-0.325$ \\
Std. Dev. & $0.063$ \\
95\% CI Lower & $-0.444$ \\
95\% CI Upper & $-0.206$ \\
$N$ replications & $50$ \\
$N$ found BP & $50$ \\
\bottomrule
\end{tabular}

\caption{Bootstrap Confidence Interval for $\delta_{crit}$ ($B=50$ replications, $\tau_{Precision}=0.7$). Each replication draws fresh samples at each drift step and identifies $\delta_{crit}$ as the first step where Precision falls below $\tau$ for three consecutive steps. The paper's reported $\delta_{crit} \approx -0.35$ (from rolling window averages) falls within the bootstrap 95\% CI.}
\label{tab:bootstrap_ci}
\end{table}

\subsection{Extended Performance Metrics}
\label{ssb:multiple-metrics}
To verify that the degradation is not an artifact of our choice of primary metric, Table~\ref{tab:multiple_metrics} reports eight complementary performance measures at each drift step: Accuracy, Balanced Accuracy, Recall, Specificity, Precision, F1, AUROC, and PR-AUC. All metrics exhibit a consistent downward trend as $|\delta|$ increases, confirming that Causal Parametric Drift produces broad classifier degradation rather than metric-specific distortion.

\begin{table}[H]
\centering
\small
\resizebox{\textwidth}{!}{%
\begin{tabular}{rrrrrrrrrr}
\toprule
$k$ & $\delta$ & Accuracy & Bal. Accuracy & Recall & Specificity & Precision & F1 & AUROC & PR-AUC \\
\midrule
0 & -0.000 & 0.789 & 0.786 & 0.833 & 0.739 & 0.783 & 0.807 & 0.847 & 0.822 \\
1 & -0.025 & 0.798 & 0.800 & 0.849 & 0.752 & 0.759 & 0.802 & 0.870 & 0.836 \\
2 & -0.050 & 0.782 & 0.779 & 0.830 & 0.728 & 0.777 & 0.802 & 0.836 & 0.824 \\
3 & -0.075 & 0.798 & 0.798 & 0.826 & 0.771 & 0.782 & 0.803 & 0.864 & 0.832 \\
4 & -0.100 & 0.803 & 0.801 & 0.838 & 0.764 & 0.801 & 0.819 & 0.870 & 0.868 \\
5 & -0.125 & 0.776 & 0.778 & 0.821 & 0.735 & 0.734 & 0.775 & 0.832 & 0.773 \\
6 & -0.150 & 0.738 & 0.742 & 0.836 & 0.648 & 0.683 & 0.752 & 0.827 & 0.794 \\
7 & -0.175 & 0.780 & 0.779 & 0.845 & 0.714 & 0.749 & 0.794 & 0.840 & 0.799 \\
8 & -0.200 & 0.774 & 0.775 & 0.822 & 0.729 & 0.739 & 0.778 & 0.837 & 0.807 \\
9 & -0.225 & 0.770 & 0.771 & 0.800 & 0.741 & 0.747 & 0.773 & 0.820 & 0.781 \\
10 & -0.250 & 0.771 & 0.777 & 0.839 & 0.714 & 0.707 & 0.767 & 0.844 & 0.793 \\
11 & -0.275 & 0.756 & 0.766 & 0.847 & 0.685 & 0.678 & 0.753 & 0.808 & 0.688 \\
12 & -0.300 & 0.734 & 0.742 & 0.789 & 0.695 & 0.644 & 0.709 & 0.802 & 0.702 \\
13 & -0.325 & 0.716 & 0.718 & 0.807 & 0.630 & 0.676 & 0.735 & 0.770 & 0.707 \\
\rowcolor{gray!20} 14 & -0.350 & 0.739 & 0.739 & 0.789 & 0.688 & 0.717 & 0.751 & 0.781 & 0.700 \\
15 & -0.375 & 0.724 & 0.729 & 0.772 & 0.686 & 0.656 & 0.709 & 0.798 & 0.696 \\
16 & -0.400 & 0.714 & 0.718 & 0.783 & 0.654 & 0.662 & 0.717 & 0.768 & 0.688 \\
17 & -0.425 & 0.695 & 0.697 & 0.762 & 0.633 & 0.657 & 0.705 & 0.752 & 0.700 \\
18 & -0.450 & 0.700 & 0.703 & 0.761 & 0.645 & 0.657 & 0.705 & 0.776 & 0.740 \\
19 & -0.475 & 0.720 & 0.721 & 0.748 & 0.694 & 0.695 & 0.721 & 0.773 & 0.722 \\
20 & -0.500 & 0.702 & 0.702 & 0.733 & 0.672 & 0.688 & 0.710 & 0.743 & 0.713 \\
\bottomrule
\end{tabular}%
}

\caption{Extended Performance Metrics Across Drift Steps in the Self Help Scenario. All metrics are computed on 500 samples generated at each step $k$ with drift magnitude $\delta$. The shaded row marks the estimated breaking point ($\delta_{crit}$).}
\label{tab:multiple_metrics}
\end{table}

\section{Comparisons with Conventional Approaches and Sensitivity Analysis}
\label{s:comparisons}
The Robustness Curve identifies a vulnerability that, by construction, leaves the input distribution untouched. This raises three natural questions: would a practitioner relying on standard monitoring tools have caught this drift; would feature-importance heuristics such as SHAP have flagged \texttt{work\_interfere} as a fragile dependency; and is the failure pattern qualitatively different from what one would obtain by simply injecting noise? We answer each in turn, then assess how much of our finding depends on the particular causal graph we discovered. All experiments in this section use the Self Help scenario (weakening the \texttt{work\_interfere} $\to$ \texttt{treatment} mechanism) with $K=20$ drift steps from $\delta=0$ to $\delta=-0.5$.

\subsection{Unsupervised Drift Monitors}
\label{ssb:unsupervised}
A central claim of our framework is that causal parametric drift evades standard unsupervised monitors. We substantiate this by deploying three families of detectors: univariate divergence (Jensen--Shannon), distributional testing (two-sample Kolmogorov--Smirnov), and multivariate reconstruction (PCA). None produces a reliable alarm despite substantial classifier degradation.

Table~\ref{tab:monitor_failure_targeted} focuses on the two features directly involved in the drifted mechanism: \texttt{work\_interfere} (manipulated parent) and \texttt{treatment} (target node). Both JS divergence and KS $p$-values are computed between reference and generated data at each drift step. JS values remain far below the alert threshold ($\tau_{JS}=0.1$), and KS $p$-values stay well above $0.05$ throughout, confirming that neither univariate test detects any distributional shift on the drifted mechanism itself.

\begin{table}[H]
\centering
\small
\begin{tabular}{rrrrrrrr}
\toprule
$k$ & $\delta$ & JS(\texttt{wi}) & KS $p$(\texttt{wi}) & JS(\texttt{trt}) & KS $p$(\texttt{trt}) & Precision & F1 \\
\midrule
0 & -0.000 & 0.004 & 0.556 & 0.001 & 0.976 & 0.783 & 0.807 \\
1 & -0.025 & 0.000 & 1.000 & 0.001 & 0.971 & 0.759 & 0.802 \\
2 & -0.050 & 0.013 & 0.396 & 0.002 & 0.951 & 0.777 & 0.802 \\
3 & -0.075 & 0.001 & 1.000 & 0.000 & 1.000 & 0.782 & 0.803 \\
4 & -0.100 & 0.001 & 0.998 & 0.001 & 0.952 & 0.801 & 0.819 \\
5 & -0.125 & 0.003 & 0.999 & 0.002 & 0.780 & 0.734 & 0.775 \\
6 & -0.150 & 0.003 & 0.872 & 0.002 & 0.902 & 0.683 & 0.752 \\
7 & -0.175 & 0.005 & 0.587 & 0.000 & 1.000 & 0.749 & 0.794 \\
8 & -0.200 & 0.007 & 0.729 & 0.001 & 0.992 & 0.739 & 0.778 \\
9 & -0.225 & 0.004 & 0.998 & 0.001 & 1.000 & 0.747 & 0.773 \\
10 & -0.250 & 0.002 & 0.990 & 0.006 & 0.232 & 0.707 & 0.767 \\
11 & -0.275 & 0.004 & 0.615 & 0.009 & 0.086 & 0.678 & 0.753 \\
12 & -0.300 & 0.008 & 0.245 & 0.018 & 0.004 & 0.644 & 0.709 \\
13 & -0.325 & 0.004 & 0.557 & 0.001 & 1.000 & 0.676 & 0.735 \\
\rowcolor{gray!20} 14 & -0.350 & 0.003 & 1.000 & 0.000 & 1.000 & 0.717 & 0.751 \\
15 & -0.375 & 0.009 & 0.832 & 0.009 & 0.071 & 0.656 & 0.709 \\
16 & -0.400 & 0.001 & 1.000 & 0.004 & 0.557 & 0.662 & 0.717 \\
17 & -0.425 & 0.002 & 0.906 & 0.001 & 0.971 & 0.657 & 0.705 \\
18 & -0.450 & 0.005 & 0.987 & 0.002 & 0.806 & 0.657 & 0.705 \\
19 & -0.475 & 0.001 & 1.000 & 0.001 & 0.992 & 0.695 & 0.721 \\
20 & -0.500 & 0.002 & 1.000 & 0.000 & 1.000 & 0.688 & 0.710 \\
\bottomrule
\end{tabular}

\caption{Univariate Drift Monitoring on the Drifted Mechanism. Jensen--Shannon divergence and two-sample Kolmogorov--Smirnov $p$-values are computed between reference and generated data for the two features directly involved in the causal drift: \texttt{work\_interfere} (\texttt{wi}, manipulated parent) and \texttt{treatment} (\texttt{trt}, target node). JS values remain far below the alert threshold ($\tau_{JS}=0.1$) and KS $p$-values stay well above $0.05$ at every step, confirming that neither univariate test detects any distributional shift on the drifted mechanism despite substantial Precision degradation. The shaded row marks the estimated breaking point ($\delta_{crit}$).}
\label{tab:monitor_failure_targeted}
\end{table}

Table~\ref{tab:monitor_failure_aggregate} broadens the analysis to all 22 input features. We report the maximum JS divergence across features (with an alert at $\tau_{JS}=0.1$), KS $p$-values for the three highest-ranked SHAP features (\texttt{work\_interfere}, \texttt{family\_history}, \texttt{care\_options}), and PCA reconstruction error (5-component model, $\mu+3\sigma$ threshold). Sporadic JS alerts are indistinguishable from sampling noise: they appear even at baseline and do not correlate with drift magnitude. KS $p$-values remain non-significant, and PCA reconstruction error never triggers. Despite the absence of any reliable alarm, Precision degrades from $0.78$ to below $0.70$.

\begin{table}[H]
\centering
\small
\resizebox{\textwidth}{!}{%
\begin{tabular}{rrrrrrrrrr}
\toprule
$k$ & $\delta$ & Max JS & JS Alert ($\tau_{JS}\!=\!0.1$) & KS $p$(\texttt{wi}) & KS $p$(\texttt{fh}) & KS $p$(\texttt{co}) & PCA RE & PCA Alert ($\mu\!+\!3\sigma$) & Precision \\
\midrule
0 & -0.000 & 0.070 & \xmark & 0.556 & 1.000 & 0.248 & 0.205 & \xmark & 0.783 \\
1 & -0.025 & 0.094 & \xmark & 1.000 & 1.000 & 0.988 & 0.206 & \xmark & 0.759 \\
2 & -0.050 & 0.150 & \cmark & 0.396 & 1.000 & 0.291 & 0.205 & \xmark & 0.777 \\
3 & -0.075 & 0.089 & \xmark & 1.000 & 0.964 & 0.973 & 0.208 & \xmark & 0.782 \\
4 & -0.100 & 0.148 & \cmark & 0.998 & 1.000 & 0.999 & 0.205 & \xmark & 0.801 \\
5 & -0.125 & 0.149 & \cmark & 0.999 & 0.998 & 1.000 & 0.208 & \xmark & 0.734 \\
6 & -0.150 & 0.128 & \cmark & 0.872 & 1.000 & 0.983 & 0.207 & \xmark & 0.683 \\
7 & -0.175 & 0.142 & \cmark & 0.587 & 0.965 & 0.999 & 0.203 & \xmark & 0.749 \\
8 & -0.200 & 0.081 & \xmark & 0.729 & 0.324 & 1.000 & 0.205 & \xmark & 0.739 \\
9 & -0.225 & 0.079 & \xmark & 0.998 & 1.000 & 1.000 & 0.207 & \xmark & 0.747 \\
10 & -0.250 & 0.131 & \cmark & 0.990 & 1.000 & 0.981 & 0.209 & \xmark & 0.707 \\
11 & -0.275 & 0.157 & \cmark & 0.615 & 1.000 & 0.748 & 0.200 & \xmark & 0.678 \\
12 & -0.300 & 0.158 & \cmark & 0.245 & 1.000 & 0.989 & 0.203 & \xmark & 0.644 \\
13 & -0.325 & 0.097 & \xmark & 0.557 & 1.000 & 1.000 & 0.206 & \xmark & 0.676 \\
\rowcolor{gray!20} 14 & -0.350 & 0.116 & \cmark & 1.000 & 1.000 & 1.000 & 0.207 & \xmark & 0.717 \\
15 & -0.375 & 0.086 & \xmark & 0.832 & 1.000 & 0.515 & 0.207 & \xmark & 0.656 \\
16 & -0.400 & 0.119 & \cmark & 1.000 & 1.000 & 0.632 & 0.208 & \xmark & 0.662 \\
17 & -0.425 & 0.149 & \cmark & 0.906 & 1.000 & 0.995 & 0.211 & \xmark & 0.657 \\
18 & -0.450 & 0.153 & \cmark & 0.987 & 0.359 & 1.000 & 0.207 & \xmark & 0.657 \\
19 & -0.475 & 0.111 & \cmark & 1.000 & 1.000 & 1.000 & 0.205 & \xmark & 0.695 \\
20 & -0.500 & 0.138 & \cmark & 1.000 & 1.000 & 0.986 & 0.207 & \xmark & 0.688 \\
\bottomrule
\end{tabular}%
}

\caption{Aggregate Unsupervised Monitoring Summary. Max JS reports the largest Jensen--Shannon divergence across all 22 input features (excluding the target). With 22 features tested per step, exceeding $\tau_{JS}=0.1$ by sampling noise alone is expected---sporadic alerts appear even at baseline ($k=0$--$4$) and do not correlate with actual drift magnitude. Two-sample KS $p$-values are shown for the three highest-ranked SHAP features: \texttt{work\_interfere} (\texttt{wi}), \texttt{family\_history} (\texttt{fh}), and \texttt{care\_options} (\texttt{co}). All $p$-values remain well above $0.05$, confirming no detectable distributional shift on the most predictively important features. PCA reconstruction error (5-component, threshold $\mu+3\sigma$ from reference data) never triggers an alert. Despite no reliable drift signal from any monitor, Precision degrades from $0.78$ to below $0.70$. The shaded row marks the estimated breaking point ($\delta_{crit}$).}
\label{tab:monitor_failure_aggregate}
\end{table}

\subsection{Supervised Detection Under Label Delay}
\label{ssb:supervised}
Even when supervised monitoring is available, its utility depends on how quickly ground-truth labels arrive. In mental-health applications, the outcome of interest (whether a patient ultimately seeks treatment) may not be observed for weeks, months, or longer. Table~\ref{tab:supervised_comparison} quantifies this limitation by simulating three label-delay scenarios.

Assuming a deployment rate of 300 observations per day and 500 samples per drift step, each step spans $\approx 1.7$ days. A supervised detector (Precision drop $>5\%$ from baseline) is evaluated under four conditions: instantaneous labels, 1-week delay (4 steps), 1-month delay (18 steps), and 1-year delay (219 steps). With instant labels, the detector fires shortly after degradation begins. With a 1-week delay, alerts lag behind by several drift steps. With a 1-month delay, the detector never fires within the 20-step simulation window, and neither does the 1-year delay, which is arguably the most realistic for mental-health outcomes. Our framework, by contrast, identifies the vulnerability pre-deployment with no label dependency whatsoever.

\begin{table}[H]
\centering
\small
\begin{tabular}{rrrrrrrrr}
\toprule
$k$ & $\delta$ & Precision & $\Delta$Prec & Instant & 1-Week & 1-Month & 1-Year & Framework \\
\midrule
0 & -0.000 & 0.783 & 0.000 & \xmark & \xmark & \xmark & \xmark & \xmark \\
1 & -0.025 & 0.759 & 0.024 & \xmark & \xmark & \xmark & \xmark & \xmark \\
2 & -0.050 & 0.777 & 0.006 & \xmark & \xmark & \xmark & \xmark & \xmark \\
3 & -0.075 & 0.782 & 0.001 & \xmark & \xmark & \xmark & \xmark & \xmark \\
4 & -0.100 & 0.801 & -0.019 & \xmark & \xmark & \xmark & \xmark & \xmark \\
5 & -0.125 & 0.734 & 0.049 & \xmark & \xmark & \xmark & \xmark & \xmark \\
\rowcolor{gray!20} 6 & -0.150 & 0.683 & 0.100 & \cmark & \xmark & \xmark & \xmark & \xmark \\
7 & -0.175 & 0.749 & 0.034 & \xmark & \xmark & \xmark & \xmark & \xmark \\
8 & -0.200 & 0.739 & 0.044 & \xmark & \xmark & \xmark & \xmark & \xmark \\
9 & -0.225 & 0.747 & 0.036 & \xmark & \xmark & \xmark & \xmark & \xmark \\
10 & -0.250 & 0.707 & 0.076 & \cmark & \cmark & \xmark & \xmark & \xmark \\
11 & -0.275 & 0.678 & 0.105 & \cmark & \xmark & \xmark & \xmark & \xmark \\
12 & -0.300 & 0.644 & 0.139 & \cmark & \xmark & \xmark & \xmark & \xmark \\
13 & -0.325 & 0.676 & 0.107 & \cmark & \xmark & \xmark & \xmark & \xmark \\
14 & -0.350 & 0.717 & 0.066 & \cmark & \cmark & \xmark & \xmark & \cmark \\
15 & -0.375 & 0.656 & 0.127 & \cmark & \cmark & \xmark & \xmark & \cmark \\
16 & -0.400 & 0.662 & 0.121 & \cmark & \cmark & \xmark & \xmark & \cmark \\
17 & -0.425 & 0.657 & 0.126 & \cmark & \cmark & \xmark & \xmark & \cmark \\
18 & -0.450 & 0.657 & 0.126 & \cmark & \cmark & \xmark & \xmark & \cmark \\
19 & -0.475 & 0.695 & 0.088 & \cmark & \cmark & \xmark & \xmark & \cmark \\
20 & -0.500 & 0.688 & 0.095 & \cmark & \cmark & \xmark & \xmark & \cmark \\
\bottomrule
\end{tabular}

\caption{Supervised Drift Detection Under Label Delay. At 300 observations/day and 500 samples per drift step, each step spans $\approx$1.7 days. A 1-week delay corresponds to 4 steps, 1-month to 18 steps, and 1-year to 219 steps. The supervised detector (Precision drop $>5\%$) fires only after delayed labels arrive; with realistic mental-health outcome delays (months to years), it never fires within the simulation window. Our framework identifies vulnerability pre-deployment with zero label dependency. The shaded row marks the first step where the supervised detector fires with instantaneous labels.}
\label{tab:supervised_comparison}
\end{table}

\subsection{Marginal Distribution Stability}
\label{ssb:marginal}
To confirm that the simulated drift is purely conceptual (a change in $P(Y \mid X)$ rather than in $P(X)$), Table~\ref{tab:marginal_stability} reports two-sample Kolmogorov--Smirnov tests between the reference data and each drifted sample for all input features. The consistently high $p$-values (well above $0.05$) demonstrate that the marginal distributions of all features remain stable throughout the simulation. This is expected by construction: modifying only the regression coefficients within a single causal mechanism alters the conditional $P(\texttt{treatment} \mid \texttt{work\_interfere})$ without perturbing the upstream distributions. The result confirms that any observed classifier degradation is attributable to concept drift alone.\footnote{An alternative perspective: in the SCM formalism, only the exogenous nodes ($U$) represent truly external factors. Our interventions modify how exogenous information propagates to $Y$ without altering $P(U)$, constituting ``pure'' concept drift by construction.}

\begin{table}[H]
\centering
\small
\resizebox{\textwidth}{!}{%
\begin{tabular}{rrrrrrrrr}
\toprule
$k$ & $\delta$ & KS(\texttt{wi}) & $p$(\texttt{wi}) & KS(\texttt{Age}) & $p$(\texttt{Age}) & Max KS (other) & Min $p$ (other) & Any $p<0.05$? \\
\midrule
0 & -0.000 & 0.042 & 0.556 & 0.063 & 0.113 & 0.054 & 0.248 & \xmark \\
1 & -0.025 & 0.003 & 1.000 & 0.072 & 0.050 & 0.051 & 0.308 & \cmark \\
2 & -0.050 & 0.047 & 0.396 & 0.055 & 0.222 & 0.052 & 0.280 & \xmark \\
3 & -0.075 & 0.016 & 1.000 & 0.066 & 0.090 & 0.046 & 0.428 & \xmark \\
4 & -0.100 & 0.020 & 0.998 & 0.055 & 0.223 & 0.051 & 0.298 & \xmark \\
5 & -0.125 & 0.020 & 0.999 & 0.074 & 0.040 & 0.068 & 0.071 & \cmark \\
6 & -0.150 & 0.031 & 0.872 & 0.082 & 0.016 & 0.072 & 0.052 & \cmark \\
7 & -0.175 & 0.041 & 0.588 & 0.061 & 0.129 & 0.041 & 0.579 & \xmark \\
8 & -0.200 & 0.036 & 0.729 & 0.063 & 0.112 & 0.050 & 0.325 & \xmark \\
9 & -0.225 & 0.021 & 0.998 & 0.056 & 0.211 & 0.034 & 0.788 & \xmark \\
10 & -0.250 & 0.023 & 0.990 & 0.066 & 0.086 & 0.027 & 0.946 & \xmark \\
11 & -0.275 & 0.040 & 0.615 & 0.064 & 0.103 & 0.055 & 0.232 & \xmark \\
12 & -0.300 & 0.054 & 0.244 & 0.076 & 0.031 & 0.038 & 0.666 & \cmark \\
13 & -0.325 & 0.042 & 0.556 & 0.055 & 0.221 & 0.087 & 0.008 & \cmark \\
\rowcolor{gray!20} 14 & -0.350 & 0.015 & 1.000 & 0.075 & 0.034 & 0.064 & 0.103 & \cmark \\
15 & -0.375 & 0.033 & 0.833 & 0.059 & 0.168 & 0.055 & 0.234 & \xmark \\
16 & -0.400 & 0.007 & 1.000 & 0.080 & 0.021 & 0.063 & 0.119 & \cmark \\
17 & -0.425 & 0.029 & 0.906 & 0.085 & 0.012 & 0.056 & 0.212 & \cmark \\
18 & -0.450 & 0.024 & 0.987 & 0.061 & 0.141 & 0.062 & 0.130 & \xmark \\
19 & -0.475 & 0.016 & 1.000 & 0.089 & 0.006 & 0.036 & 0.721 & \cmark \\
20 & -0.500 & 0.016 & 1.000 & 0.071 & 0.055 & 0.037 & 0.692 & \xmark \\
\bottomrule
\end{tabular}%
}

\caption{Marginal Distribution Stability: Two-sample Kolmogorov-Smirnov test for $P(X)$ at each drift step. With 22 features $\times$ 21 steps = 462 tests, sporadic $p<0.05$ values are expected by chance (approximately 5\%). The \texttt{work\_interfere} marginal remains consistently non-significant, confirming pure concept drift ($P(Y|X)$ changes while $P(X)$ remains stable). The shaded row marks the estimated breaking point ($\delta_{crit}$).}
\label{tab:marginal_stability}
\end{table}

\subsection{SHAP Feature Attribution vs. Causal Distance}
\label{ssb:shap}
Practitioners often rely on SHAP to rank features by predictive importance and decide which inputs deserve close monitoring. Table~\ref{tab:shap_comparison} augments this correlational ranking with each feature's causal distance to the target node in the learned SCM. To move beyond a single drift scenario, we ran Causal Parametric Drift Simulation on the top-5 SHAP-ranked features. For each feature, we identified the causal mechanism linking it to \texttt{treatment} (directly for distance-1 parents, or via an intermediate node for distance-2 features) and progressively weakened the relationship. The $\delta_{crit}$ column reports the breaking point for each; \textit{robust} indicates that no breaking point was found within the simulation range ($\delta_{max}=-0.5$). The result demonstrates that not all high-importance features induce classifier failure when their causal mechanisms drift, a distinction invisible to correlational tools like SHAP.

\begin{table}[H]
\centering
\small
\begin{tabular}{rrrrr}
\toprule
Feature & Mean |SHAP| & Causal Dist. & Drift Tested? & $\delta_{crit}$ \\
\midrule
\texttt{work\_interfere} & 2.0634 & 1 & \cmark & $-0.28$ \\
\texttt{family\_history} & 0.3387 & 1 & \cmark & robust \\
\texttt{care\_options} & 0.2877 & 1 & \cmark & robust \\
\texttt{benefits} & 0.2785 & 1 & \cmark & robust \\
\texttt{seek\_help} & 0.2274 & 2 & \cmark & robust \\
\texttt{Age} & 0.2213 & 2 & -- & -- \\
\texttt{Gender} & 0.1929 & 1 & -- & -- \\
\texttt{leave} & 0.1390 & 2 & -- & -- \\
\texttt{phys\_health\_interview} & 0.1308 & 2 & -- & -- \\
\texttt{coworkers} & 0.1287 & 1 & -- & -- \\
\texttt{wellness\_program} & 0.1252 & 2 & -- & -- \\
\texttt{no\_employees} & 0.1233 & $\infty$ & -- & -- \\
\texttt{mental\_health\_consequence} & 0.1127 & 2 & -- & -- \\
\texttt{anonymity} & 0.0954 & 2 & -- & -- \\
\texttt{obs\_consequence} & 0.0749 & 3 & -- & -- \\
\texttt{tech\_company} & 0.0579 & $\infty$ & -- & -- \\
\texttt{mental\_vs\_physical} & 0.0499 & 3 & -- & -- \\
\texttt{supervisor} & 0.0423 & $\infty$ & -- & -- \\
\texttt{remote\_work} & 0.0415 & $\infty$ & -- & -- \\
\texttt{phys\_health\_consequence} & 0.0394 & 3 & -- & -- \\
\texttt{mental\_health\_interview} & 0.0311 & $\infty$ & -- & -- \\
\texttt{self\_employed} & 0.0153 & 3 & -- & -- \\
\bottomrule
\end{tabular}

\caption{SHAP Feature Attribution vs.\ Causal Distance to Target. Drift simulations were run on the top 5 SHAP-ranked features ($K=20$ steps, $\delta_{max}=-0.5$). $\delta_{crit}$ is the first $\delta$ at which Precision falls below $\tau_{Precision}=0.7$ for 3 consecutive steps; \textit{robust} indicates no breaking point was found within the simulation range.}
\label{tab:shap_comparison}
\end{table}

\subsection{Causal Drift vs. Replacement Noise}
\label{ssb:replacement}
One might ask whether similar degradation patterns could be obtained by simply injecting random noise into the key feature. Table~\ref{tab:replacement_noise} compares our causally-informed drift against a na\"ive replacement-noise baseline that progressively shuffles \texttt{work\_interfere} values (random permutation), breaking all correlations with other variables while preserving the marginal distribution. Both approaches start from the same GCM-generated baseline.

The $\Delta$Prec column (causal minus noise Precision) reveals the distinction: at moderate drift levels, the causal approach degrades the classifier \textit{less} than noise, because it preserves all structural dependencies except the targeted mechanism. At higher levels, replacement noise degrades faster precisely because it destroys correlations indiscriminately. This confirms that causal drift produces a qualitatively different (and more realistic) failure mode than feature-level perturbation.

\begin{table}[H]
\centering
\small
\begin{tabular}{rrrrrrr}
\toprule
$k$ & Fraction & Causal Prec & Causal F1 & Noise Prec & Noise F1 & $\Delta$Prec \\
\midrule
0 & 0.000 & 0.783 & 0.807 & 0.783 & 0.807 & 0.000 \\
1 & 0.050 & 0.759 & 0.802 & 0.777 & 0.797 & -0.018 \\
2 & 0.100 & 0.777 & 0.802 & 0.754 & 0.776 & 0.023 \\
3 & 0.150 & 0.782 & 0.803 & 0.741 & 0.766 & 0.040 \\
4 & 0.200 & 0.801 & 0.819 & 0.734 & 0.744 & 0.067 \\
5 & 0.250 & 0.734 & 0.775 & 0.723 & 0.750 & 0.011 \\
6 & 0.300 & 0.683 & 0.752 & 0.704 & 0.742 & -0.021 \\
7 & 0.350 & 0.749 & 0.794 & 0.709 & 0.714 & 0.040 \\
8 & 0.400 & 0.739 & 0.778 & 0.689 & 0.713 & 0.050 \\
9 & 0.450 & 0.747 & 0.773 & 0.687 & 0.708 & 0.060 \\
10 & 0.500 & 0.707 & 0.767 & 0.701 & 0.707 & 0.005 \\
11 & 0.550 & 0.678 & 0.753 & 0.643 & 0.658 & 0.035 \\
12 & 0.600 & 0.644 & 0.709 & 0.635 & 0.675 & 0.009 \\
13 & 0.650 & 0.676 & 0.735 & 0.641 & 0.656 & 0.035 \\
\rowcolor{gray!20} 14 & 0.700 & 0.717 & 0.751 & 0.650 & 0.676 & 0.067 \\
15 & 0.750 & 0.656 & 0.709 & 0.617 & 0.642 & 0.039 \\
16 & 0.800 & 0.662 & 0.717 & 0.614 & 0.623 & 0.048 \\
17 & 0.850 & 0.657 & 0.705 & 0.580 & 0.604 & 0.077 \\
18 & 0.900 & 0.657 & 0.705 & 0.549 & 0.559 & 0.107 \\
19 & 0.950 & 0.695 & 0.721 & 0.577 & 0.599 & 0.118 \\
20 & 1.000 & 0.688 & 0.710 & 0.545 & 0.560 & 0.143 \\
\bottomrule
\end{tabular}

\caption{Causal Parametric Drift vs.\ Replacement Noise. Both approaches start from the same GCM-generated baseline. Causal drift modifies the \texttt{work\_interfere} $\to$ \texttt{treatment} mechanism (preserving causal structure), while replacement noise shuffles \texttt{work\_interfere} values (breaking correlations). The $\Delta$Prec column shows the difference: negative values indicate causal drift degrades more than noise, exposing a genuine vulnerability rather than an artifact of broken dependencies. The shaded row marks the estimated breaking point ($\delta_{crit}$).}
\label{tab:replacement_noise}
\end{table}

\subsection{Sensitivity to Structural Assumptions}
\label{ssb:sensitivity}
\begin{figure}[H]
    \centering
\includegraphics[width=0.9\linewidth]{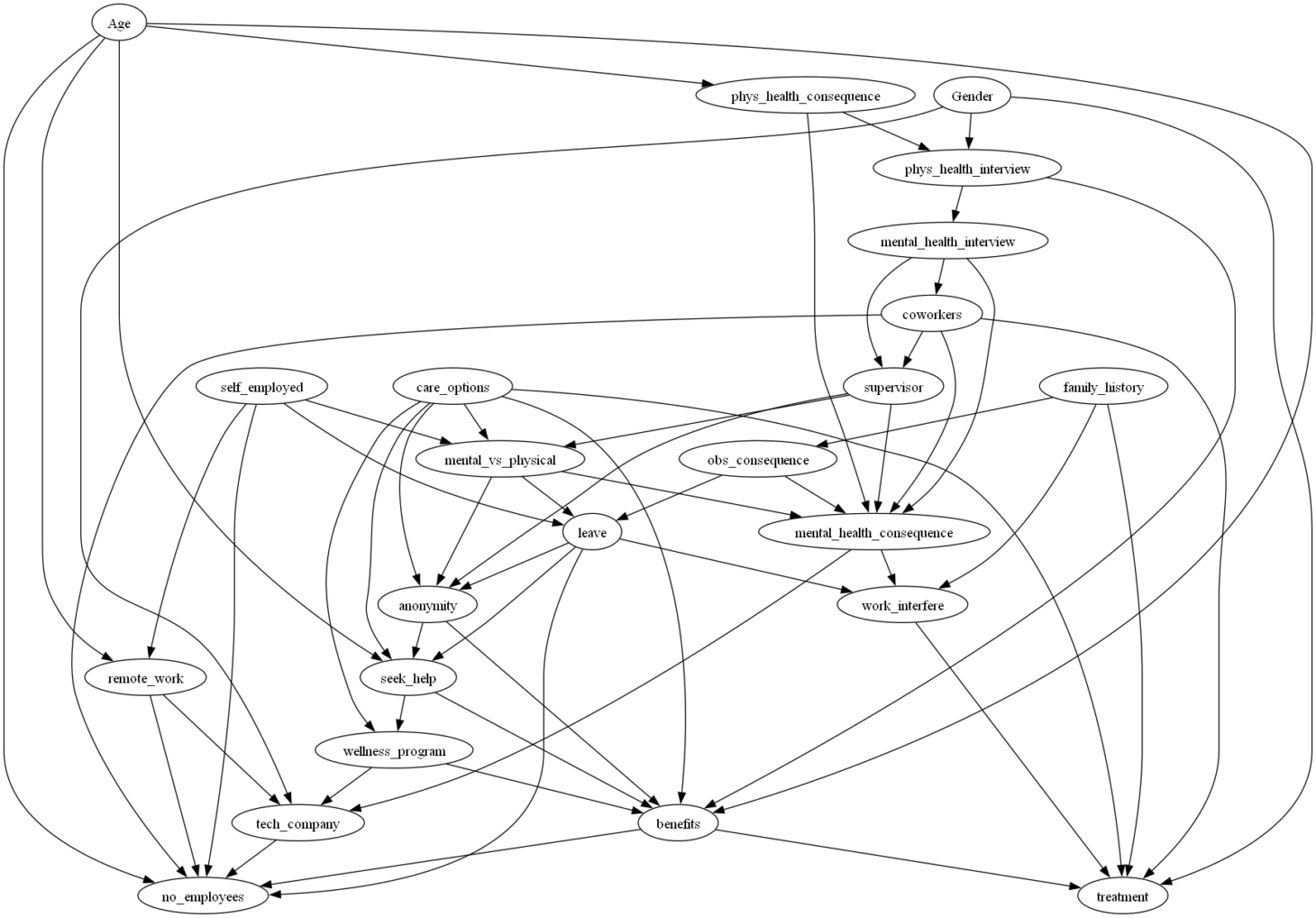}
    \caption{Variant~A: causal graph topology obtained with $\alpha_{PC}=0.1$ and partial domain constraints.}
    \label{fig:osmh_topology_alpha}
\end{figure}

An obvious concern with any framework built on causal discovery is the \textit{Rashomon Set}: the discovered SCM is one of many graphs that fit the data well, and a different graph could lead to a different Breaking Point \citep{semenova_existence_2022}. To assess the stability of our findings against this uncertainty, we conducted a sensitivity analysis on the Self Help scenario. The original Digital Twin was constructed with $\alpha_{PC}=0.05$ and strict domain constraints (e.g., enforcing \textit{Age} and \textit{Gender} as exogenous). For this analysis, we generated three alternative causal graphs:
\begin{itemize}
    \item \textbf{Variant~A:} $\alpha_{PC}=0.10$, with only \texttt{Age} and \texttt{Gender} forced exogenous (Figure~\ref{fig:osmh_topology_alpha});
    \item \textbf{Variant~B:} $\alpha_{PC}=0.20$, no domain constraints (Figure~\ref{fig:osmh_topology_alpha-0.2});
    \item \textbf{Variant~C (anti-causal):} $\alpha_{PC}=0.20$, with the deliberately implausible constraint $\texttt{treatment} \to \texttt{benefits}$ (Figure~\ref{fig:osmh_topology_alpha-0.2_ac}).
\end{itemize}

We re-fitted the SCM parameters for each new structure and re-executed the Self Help parametric drift simulation. As shown in Table~\ref{tab:sensitivity_table}, despite the resulting variations in graph topology (including the emergence of spurious edges between previously constrained variables), the Breaking Point ($\delta_{crit}$) for the XGBoost classifier remained near $-0.35$ for $\tau_{Precision}=0.7$. This shows that the identified vulnerability is robust to reasonable variations in the discovered causal structure and depends on the classifier rather than on the specifics of the data-generation mechanism.

\begin{table}[h]
\centering
\begin{tabular}{lccc}
\toprule
\textbf{Model Configuration} & \textbf{Alpha ($\alpha_{PC}$)} & \textbf{Domain Constraints} & \textbf{Breaking Point ($\delta_{crit}$)} \\
\midrule
Original Model (Fig.~\ref{fig:xgboost_robustness_curve}) & $0.05$ & All & $\approx -0.35$ \\
Variant~A (Fig.~\ref{fig:robust-curve-alpha-0.1}) & $0.10$ & [\texttt{Age}, \texttt{Gender}] & $\approx -0.35$ \\
Variant~B (Fig.~\ref{fig:osmh_topology_alpha-0.2}) & $0.20$ & None & $\approx -0.30$ \\
Variant~C, anti-causal (Fig.~\ref{fig:osmh_topology_alpha-0.2_ac}) & $0.20$ & $\texttt{treatment} \to \texttt{benefits}$ & $\approx -0.40$ \\
\bottomrule
\end{tabular}
\caption{Breaking Point ($\delta_{crit}$) across different hyperparameters of the PC algorithm during structure learning.}
\label{tab:sensitivity_table}
\end{table}

\begin{figure}[H]
    \centering
\includegraphics[width=1\linewidth]{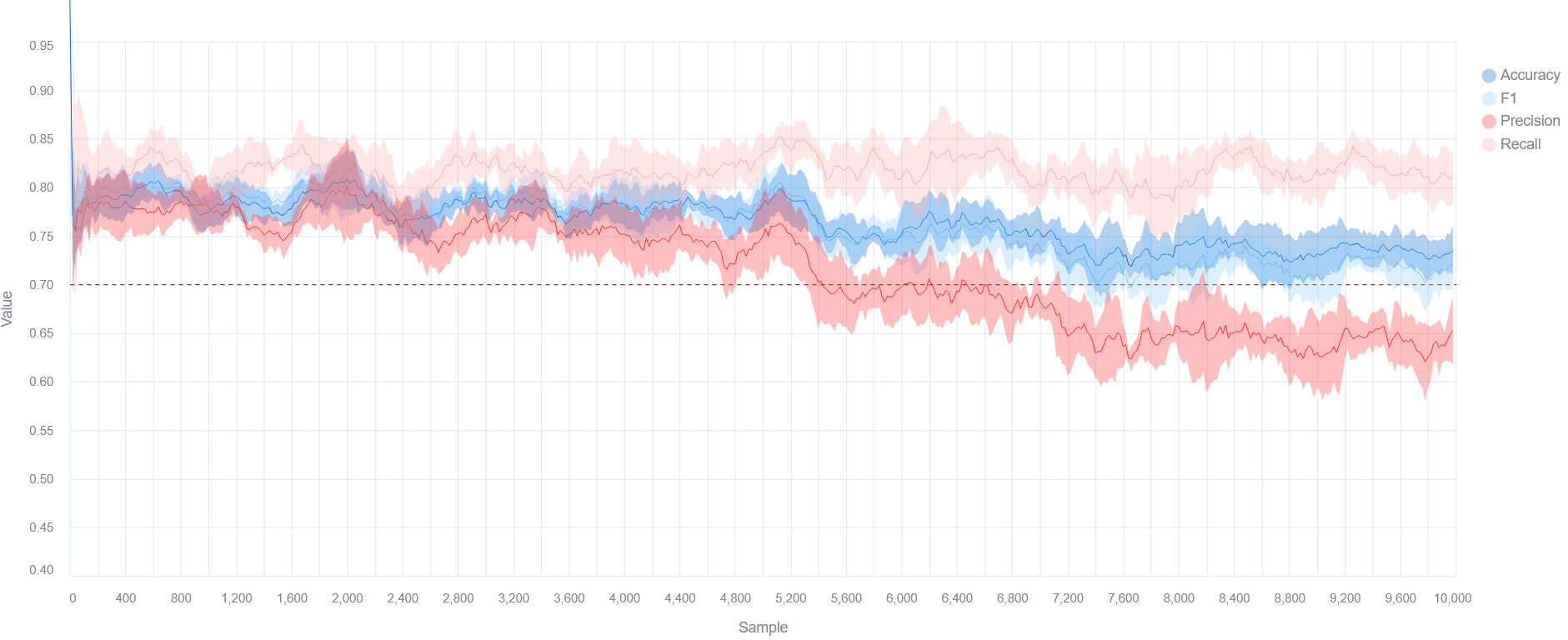}
\vspace{-0.55cm}
\caption{XGBoost Robustness Curve under the Self Help scenario, using the Variant~A topology ($\alpha_{PC}=0.1$, partial constraints). $\mathcal{D}_0$ for the first 3000 observations, then $\mathcal{D}_1$ to $\mathcal{D}_{k-1}$ for 4000 observations, then $\mathcal{D}_K$ for 3000 observations. The red dashed line marks $\tau_{Precision}=0.7$. Variant~A is shown here as an illustrative example; the topologies and robustness curves for Variants~B and~C are deferred to Appendix~\ref{a:sensitivity-figures} (Figures~\ref{fig:osmh_topology_alpha-0.2}--\ref{fig:anti-causal-robust-curve-alpha-0.2}).}
    \label{fig:robust-curve-alpha-0.1}
\end{figure}

\section{Discussion}

\paragraph{From Passive Monitoring to Active Stress-Testing.}
Current drift detection methods rely on passive statistical monitoring that often misses complex multidimensional changes \citep{vangala2022mlops, jourdan2024addressing}. Our framework enables \textit{active diagnostics}: by simulating specific parametric shifts, practitioners can stress-test models against concrete causal hypotheses, anticipating failure modes before deployment rather than analyzing them post-hoc.

\paragraph{Quantifiable Safety Margins.}
The Self Help scenario demonstrates the ability to identify a model's precise Breaking Point. Unlike correlational explainability tools that provide local feature attributions, our framework quantifies how much a specific causal mechanism must shift to degrade performance below a safety threshold. This exposes latent vulnerabilities, such as a model's over-reliance on certain features, that standard validation-set checks fail to detect. Effective scenario design requires collaboration between ML practitioners and domain experts; even implausible scenarios can reveal informative sensitivities worth investigating. Bootstrap resampling (Section~\ref{ssb:bootstrap-ci}) yields $\delta_{crit} \in [-0.44, -0.21]$ (95\% CI, $B=50$), reinforcing its use as an order-of-magnitude estimate rather than a precise threshold. The SHAP comparison in Section~\ref{ssb:shap} further shows that not all high-attribution features induce failure when their causal mechanisms drift.

\paragraph{Informed Monitoring Strategies.}
Our framework complements rather than replaces statistical drift detection. While statistical monitors excel at detecting distributional shifts in $P(X)$, they cannot anticipate vulnerabilities to concept drift in $P(Y|X)$. The Digital Twin enables preemptive identification of which relationships to monitor, which thresholds matter, and what contingency plans to prepare: insights that statistical monitoring alone cannot provide. After deployment, practitioners should combine statistical monitoring (for real-time anomaly detection) with the causal insights from stress-testing (for targeted, informed alerting). Section~\ref{s:comparisons} substantiates this empirically: unsupervised monitors raise no alarm at the Breaking Point (Tables~\ref{tab:monitor_failure_targeted}--\ref{tab:monitor_failure_aggregate}), supervised detectors are delayed by realistic label lag (Table~\ref{tab:supervised_comparison}), and marginal distributions remain stable throughout (Table~\ref{tab:marginal_stability}).

\paragraph{A Continuous Diagnostic Sandbox.}
The Digital Twin acts as a living artifact that can be reapplied to new data throughout the production lifecycle, comparing the evolving causal structure against initial hypotheses. This white-box view enables confident decisions regarding monitoring adaptation or model retirement, while invalidating outdated assumptions as ground truth becomes available.

\paragraph{The Rashomon Set.}
A limitation of our approach is that the discovered SCM represents \textit{one} possible data-generating mechanism, not necessarily the \textit{true} one, a fact known as the \textit{Rashomon Set} \citep{semenova_existence_2022}. Section~\ref{ssb:sensitivity} shows that Causal Parametric Drift Simulation is robust to shifts in topology and Markov-equivalence classes; nonetheless, we recommend treating Digital Twin robustness estimates as a lower bound for specific causal hypotheses and as useful discussion tools rather than a perfect representation of the data-generating mechanism.

\paragraph{Data Constraints and Scalability.}
The fidelity of the Digital Twin relies heavily on the quality of the observational data and of the causal-discovery algorithm. By trying to maximize the interpretability of the generation mechanisms, we restrict the form that $f_v$ can take. As such, the framework is currently limited in cases where linearity does not hold. Additionally, high-dimensional datasets may yield dense graphs that are difficult to interpret and manipulate. Causal drift also produces qualitatively different failure modes than na\"ive feature replacement (Section~\ref{ssb:replacement}), and degradation persists across all standard metrics (Section~\ref{ssb:multiple-metrics}).

\section{Conclusion}

Our framework leverages interpretable Structural Causal Models to create Digital Twins of the data-generation and classification process and which allow practitioners to stress-test pipelines against specific mechanisms and to prepare contingency plans rather than reacting ad-hoc to opaque alerts.  Current industry standards often rely on statistical monitoring by tracking divergences in marginal or joint distributions which frequently fails to distinguish between benign covariate shifts and genuine structural decay. By contrast, our approach exposes latent vulnerabilities and quantifiable breaking points. Ultimately, by anchoring evaluation in the structural invariants of the environment rather than transient statistics, this framework transforms drift analysis from a reactive alarm system into a robust, anticipatory validation tool for high-stakes environments.

Future research should relax the linearity constraint that currently restricts the framework, by interpreting parameter modifications in non-linear models (e.g., additive noise models) as nodal data-generation mechanisms in the SCM. Developing rigorous fit metrics for SCMs beyond global covariance alignment would also help mitigate the implications of the \textit{Rashomon Set}. We further aim to enhance post-hoc explainability through this causal environment: randomizing structural parameters rather than feature values yields feature importances that respect the causal structure. Applying the framework to a broader range of tasks would help establish causal simulation as a standard practice for deploying trustworthy Machine Learning systems in high-stakes domains. We advocate for a paradigm shift in drift monitoring---from passive statistical observation to active, scenario-based causal diagnostics.

\newpage
\bibliography{Maitrise}

\begin{thebibliography}{58}
\providecommand{\natexlab}[1]{#1}
\providecommand{\url}[1]{\texttt{#1}}
\expandafter\ifx\csname urlstyle\endcsname\relax
  \providecommand{\doi}[1]{doi: #1}\else
  \providecommand{\doi}{doi: \begingroup \urlstyle{rm}\Url}\fi

\bibitem[{Abdul Razak} et~al.(2023){Abdul Razak}, Nirmala, Sreenivasa, Lahza,
  and Lahza]{m_s_survey_2023}
M.~S. {Abdul Razak}, C.~R. Nirmala, B.~R. Sreenivasa, Husam Lahza, and Hassan
  Fareed~M. Lahza.
\newblock A survey on detecting healthcare concept drift in {AI}/{ML} models
  from a finance perspective.
\newblock \emph{Frontiers in Artificial Intelligence}, 5, April 2023.
\newblock ISSN 2624-8212.
\newblock \doi{10.3389/frai.2022.955314}.
\newblock Publisher: Frontiers.

\bibitem[Arjovsky et~al.(2019)Arjovsky, Bottou, Gulrajani, and
  Lopez-Paz]{arjovsky2019invariant}
Martin Arjovsky, L{\'e}on Bottou, Ishaan Gulrajani, and David Lopez-Paz.
\newblock Invariant risk minimization.
\newblock \emph{arXiv preprint arXiv:1907.02893}, 2019.

\bibitem[Ashmore et~al.(2022)Ashmore, Calinescu, and
  Paterson]{ashmore_assuring_2022}
Rob Ashmore, Radu Calinescu, and Colin Paterson.
\newblock Assuring the {Machine} {Learning} {Lifecycle}: {Desiderata},
  {Methods}, and {Challenges}.
\newblock \emph{ACM Computing Surveys}, 54\penalty0 (5):\penalty0 1--39, June
  2022.
\newblock ISSN 0360-0300, 1557-7341.
\newblock \doi{10.1145/3453444}.

\bibitem[Bl{{\"o}}baum et~al.(2024)Bl{{\"o}}baum, G{{\"o}}tz, Budhathoki,
  Mastakouri, and Janzing]{dowhy_blaum}
Patrick Bl{{\"o}}baum, Peter G{{\"o}}tz, Kailash Budhathoki, Atalanti~A.
  Mastakouri, and Dominik Janzing.
\newblock Dowhy-gcm: An extension of dowhy for causal inference in graphical
  causal models.
\newblock \emph{Journal of Machine Learning Research}, 25\penalty0
  (147):\penalty0 1--7, 2024.
\newblock URL \url{http://jmlr.org/papers/v25/22-1258.html}.

\bibitem[Chen et~al.(2022)Chen, Zaharia, and Zou]{chen_estimating_2022}
Lingjiao Chen, Matei Zaharia, and James Zou.
\newblock Estimating and {Explaining} {Model} {Performance} {When} {Both}
  {Covariates} and {Labels} {Shift}.
\newblock 2022.
\newblock \doi{10.48550/ARXIV.2209.08436}.

\bibitem[Chen and Guestrin(2016)]{xgboost_2016}
Tianqi Chen and Carlos Guestrin.
\newblock Xgboost: A scalable tree boosting system.
\newblock In \emph{Proceedings of the 22nd ACM SIGKDD International Conference
  on Knowledge Discovery and Data Mining}, KDD '16, page 785–794, New York,
  NY, USA, 2016. Association for Computing Machinery.
\newblock ISBN 9781450342322.
\newblock \doi{10.1145/2939672.2939785}.
\newblock URL \url{https://doi.org/10.1145/2939672.2939785}.

\bibitem[Chuah et~al.(2022)Chuah, Kruger, Wang, Yan, and
  Hahn]{chuah_framework_2022}
Joshua Chuah, Uwe Kruger, Ge~Wang, Pingkun Yan, and Juergen Hahn.
\newblock Framework for {Testing} {Robustness} of {Machine} {Learning}-{Based}
  {Classifiers}.
\newblock \emph{Journal of Personalized Medicine}, 12\penalty0 (8):\penalty0
  1314, August 2022.
\newblock ISSN 2075-4426.
\newblock \doi{10.3390/jpm12081314}.
\newblock Number: 8 Publisher: Multidisciplinary Digital Publishing Institute.

\bibitem[Cohen(1988)]{cohen1988statistical}
Jacob Cohen.
\newblock \emph{Statistical Power Analysis for the Behavioral Sciences}.
\newblock Lawrence Erlbaum Associates, 2nd edition, 1988.

\bibitem[Daza et~al.(2020)Daza, Castillo, Escobar, Valencia, Pinz\'{o}n, and
  Arbel\'{a}ez]{lucas_dataset}
Laura Daza, Angela Castillo, Mar\'{\i}a Escobar, Sergio Valencia, Bibiana
  Pinz\'{o}n, and Pablo Arbel\'{a}ez.
\newblock Lucas: Lung cancer screening with multimodal biomarkers.
\newblock In \emph{Multimodal Learning for Clinical Decision Support and
  Clinical Image-Based Procedures: 10th International Workshop, ML-CDS 2020,
  and 9th International Workshop, CLIP 2020, Held in Conjunction with MICCAI
  2020, Lima, Peru, October 4–8, 2020, Proceedings}, page 115–124, Berlin,
  Heidelberg, 2020. Springer-Verlag.
\newblock ISBN 978-3-030-60945-0.
\newblock \doi{10.1007/978-3-030-60946-7_12}.
\newblock URL
  \url{https://doi-org.acces.bibl.ulaval.ca/10.1007/978-3-030-60946-7_12}.

\bibitem[Dodge(2008)]{kolmo_smirnov}
Yadolah Dodge.
\newblock \emph{Kolmogorov--Smirnov Test}, pages 283--287.
\newblock Springer New York, New York, NY, 2008.
\newblock ISBN 978-0-387-32833-1.
\newblock \doi{10.1007/978-0-387-32833-1_214}.
\newblock URL \url{https://doi.org/10.1007/978-0-387-32833-1_214}.

\bibitem[Frye et~al.(2020)Frye, Rowat, and Feige]{frye2020asymmetric}
Christopher Frye, Colin Rowat, and Ilya Feige.
\newblock Asymmetric shapley values: incorporating causal knowledge into
  model-agnostic explainability.
\newblock \emph{Advances in neural information processing systems},
  33:\penalty0 1229--1239, 2020.

\bibitem[Gemaque et~al.(2020)Gemaque, Costa, Giusti, and
  Dos~Santos]{gemaque2020overview}
Rosana~Noronha Gemaque, Albert Fran{\c{c}}a~Josu{\'a} Costa, Rafael Giusti, and
  Eulanda~Miranda Dos~Santos.
\newblock An overview of unsupervised drift detection methods.
\newblock \emph{Wiley Interdisciplinary Reviews: Data Mining and Knowledge
  Discovery}, 10\penalty0 (6):\penalty0 e1381, 2020.

\bibitem[Glymour et~al.(2019)Glymour, Zhang, and Spirtes]{glymour_review_2019}
Clark Glymour, Kun Zhang, and Peter Spirtes.
\newblock Review of {Causal} {Discovery} {Methods} {Based} on {Graphical}
  {Models}.
\newblock \emph{Frontiers in Genetics}, 10, June 2019.
\newblock ISSN 1664-8021.
\newblock \doi{10.3389/fgene.2019.00524}.
\newblock Publisher: Frontiers.

\bibitem[Gonçalves et~al.(2014)Gonçalves, De~Carvalho~Santos, Barros, and
  Vieira]{goncalves_comparative_2014}
Paulo~M. Gonçalves, Silas~G.T. De~Carvalho~Santos, Roberto~S.M. Barros, and
  Davi~C.L. Vieira.
\newblock A comparative study on concept drift detectors.
\newblock \emph{Expert Systems with Applications}, 41\penalty0 (18):\penalty0
  8144--8156, December 2014.
\newblock ISSN 09574174.
\newblock \doi{10.1016/j.eswa.2014.07.019}.

\bibitem[Goodfellow et~al.(2014)Goodfellow, Pouget-Abadie, Mirza, Xu,
  Warde-Farley, Ozair, Courville, and Bengio]{Goodfellow2014Generative}
Ian~J. Goodfellow, Jean Pouget-Abadie, Mehdi Mirza, Bing Xu, David
  Warde-Farley, Sherjil Ozair, Aaron~C. Courville, and Yoshua Bengio.
\newblock Generative adversarial nets.
\newblock In \emph{Neural Information Processing Systems}, 2014.
\newblock URL \url{https://api.semanticscholar.org/CorpusID:261560300}.

\bibitem[Grassi et~al.(2022)Grassi, Palluzzi, and Tarantino]{SEMgraph}
Mario Grassi, Fernando Palluzzi, and Barbara Tarantino.
\newblock Semgraph: an r package for causal network inference of
  high-throughput data with structural equation models.
\newblock \emph{Bioinformatics}, 38\penalty0 (20):\penalty0 4829--4830, 08
  2022.
\newblock ISSN 1367-4811.
\newblock \doi{10.1093/bioinformatics/btac567}.
\newblock URL \url{https://doi.org/10.1093/bioinformatics/btac567}.

\bibitem[Grieves(2023)]{grieves2023digital}
Michael~W Grieves.
\newblock Digital twins: past, present, and future.
\newblock In \emph{The digital twin}, pages 97--121. Springer, 2023.

\bibitem[Hashmani et~al.(2020)Hashmani, Jameel, Rehman, and
  Inoue]{hashmani_concept_2020}
Manzoor~Ahmed Hashmani, Syed~Muslim Jameel, Mobashar Rehman, and Atsushi Inoue.
\newblock Concept {Drift} {Evolution} {In} {Machine} {Learning} {Approaches}:
  {A} {Systematic} {Literature} {Review}.
\newblock \emph{International Journal on Smart Sensing and Intelligent
  Systems}, 13\penalty0 (1):\penalty0 1--16, January 2020.
\newblock \doi{10.21307/ijssis-2020-029}.

\bibitem[Hern{\'a}n and Robins(2020)]{hernan2020causal}
Miguel~A Hern{\'a}n and James~M Robins.
\newblock \emph{Causal Inference: What If}.
\newblock Chapman \& Hall/CRC, 2020.

\bibitem[Hooper et~al.(2008)Hooper, Coughlan, and
  Mullen]{Hooper2008StructuralEM}
Daire Hooper, Joseph Coughlan, and Michael~R. Mullen.
\newblock Structural equation modelling: guidelines for determining model fit.
\newblock 2008.
\newblock URL \url{https://api.semanticscholar.org/CorpusID:32672489}.

\bibitem[Janzing et~al.(2020)Janzing, Minorics, and
  Bl{\"o}baum]{janzing2020feature}
Dominik Janzing, Lenon Minorics, and Patrick Bl{\"o}baum.
\newblock Feature relevance quantification in explainable ai: A causal problem.
\newblock In \emph{International Conference on artificial intelligence and
  statistics}, pages 2907--2916. PMLR, 2020.

\bibitem[Jourdan(2024)]{jourdan2024addressing}
Nicolas Jourdan.
\newblock Addressing concept drift in machine learning-based monitoring of
  manufacturing processes.
\newblock 2024.

\bibitem[Kingma and Welling(2013)]{Kingma2013Auto}
Diederik~P. Kingma and Max Welling.
\newblock Auto-encoding variational bayes.
\newblock \emph{CoRR}, abs/1312.6114, 2013.
\newblock URL \url{https://api.semanticscholar.org/CorpusID:216078090}.

\bibitem[Kocaoglu et~al.(2018)Kocaoglu, Snyder, Dimakis, and
  Vishwanath]{kocaoglu2018causalgan}
Murat Kocaoglu, Christopher Snyder, Alexandros~G. Dimakis, and Sriram
  Vishwanath.
\newblock {CausalGAN}: Learning causal implicit generative models with
  adversarial training.
\newblock In \emph{International Conference on Learning Representations}, 2018.

\bibitem[Kumar et~al.(2020)Kumar, Venkatasubramanian, Scheidegger, and
  Friedler]{Kumar2020Problems}
Indra~Elizabeth Kumar, Suresh Venkatasubramanian, Carlos~Eduardo Scheidegger,
  and Sorelle~A. Friedler.
\newblock Problems with shapley-value-based explanations as feature importance
  measures.
\newblock In \emph{International Conference on Machine Learning}, 2020.
\newblock URL \url{https://api.semanticscholar.org/CorpusID:211296386}.

\bibitem[Lu et~al.(2018)Lu, Liu, Dong, Gu, Gama, and Zhang]{lu_learning_2018}
Jie Lu, Anjin Liu, Fan Dong, Feng Gu, Joao Gama, and Guangquan Zhang.
\newblock Learning under {Concept} {Drift}: {A} {Review}.
\newblock \emph{IEEE Transactions on Knowledge and Data Engineering}, 2018.
\newblock ISSN 1041-4347, 1558-2191, 2326-3865.
\newblock \doi{10.1109/TKDE.2018.2876857}.
\newblock arXiv:2004.05785 [cs, stat].

\bibitem[Lundberg and Lee(2017)]{lundberg_unified_2017}
Scott~M. Lundberg and Su-In Lee.
\newblock A {Unified} {Approach} to {Interpreting} {Model} {Predictions}.
\newblock May 2017.

\bibitem[McHugh(2013)]{chi-square-cramer-v}
Mary McHugh.
\newblock The chi-square test of independence.
\newblock \emph{Biochemia medica}, 23:\penalty0 143--9, 06 2013.
\newblock \doi{10.11613/BM.2013.018}.

\bibitem[Mihai et~al.(2022)Mihai, Yaqoob, Hung, Davis, Towakel, Raza,
  Karamanoglu, Barn, Shetve, Prasad, Venkataraman, Trestian, and
  Nguyen]{mihai_digital_2022}
Stefan Mihai, Mahnoor Yaqoob, Dang~V. Hung, William Davis, Praveer Towakel,
  Mohsin Raza, Mehmet Karamanoglu, Balbir Barn, Dattaprasad Shetve, Raja~V.
  Prasad, Hrishikesh Venkataraman, Ramona Trestian, and Huan~X. Nguyen.
\newblock Digital {Twins}: {A} {Survey} on {Enabling} {Technologies},
  {Challenges}, {Trends} and {Future} {Prospects}.
\newblock \emph{IEEE Communications Surveys \& Tutorials}, 24\penalty0
  (4):\penalty0 2255--2291, 2022.
\newblock ISSN 1553-877X.
\newblock \doi{10.1109/COMST.2022.3208773}.

\bibitem[Nielsen and Cortina(2025)]{nielsen_calculating_2025}
Bo~Bernhard Nielsen and Jose~M. Cortina.
\newblock Calculating and reporting degrees of freedom in structural equation
  modeling: an empirical generalization study.
\newblock \emph{Journal of International Business Studies}, April 2025.
\newblock ISSN 1478-6990.
\newblock \doi{10.1057/s41267-025-00781-3}.
\newblock URL \url{https://doi.org/10.1057/s41267-025-00781-3}.

\bibitem[Pawlowski et~al.(2020)Pawlowski, Castro, and
  Glocker]{pawlowski2020deep}
Nick Pawlowski, Daniel~C. Castro, and Ben Glocker.
\newblock Deep structural causal models for tractable counterfactual inference.
\newblock In \emph{Advances in Neural Information Processing Systems}, 2020.

\bibitem[Pearl(2009{\natexlab{a}})]{pearl2009causal}
Judea Pearl.
\newblock Causal inference in statistics: An overview.
\newblock 2009{\natexlab{a}}.

\bibitem[Pearl(2009{\natexlab{b}})]{pearl_causality_2009}
Judea Pearl.
\newblock \emph{Causality: {Models}, {Reasoning} and {Inference}}.
\newblock Cambridge University Press, USA, 2nd edition, August
  2009{\natexlab{b}}.
\newblock ISBN 978-0-521-89560-6.

\bibitem[Pedregosa et~al.(2011)Pedregosa, Varoquaux, Gramfort, Michel, Thirion,
  Grisel, Blondel, Prettenhofer, Weiss, Dubourg, Vanderplas, Passos,
  Cournapeau, Brucher, Perrot, and Duchesnay]{scikit-learn}
F.~Pedregosa, G.~Varoquaux, A.~Gramfort, V.~Michel, B.~Thirion, O.~Grisel,
  M.~Blondel, P.~Prettenhofer, R.~Weiss, V.~Dubourg, J.~Vanderplas, A.~Passos,
  D.~Cournapeau, M.~Brucher, M.~Perrot, and E.~Duchesnay.
\newblock Scikit-learn: Machine learning in {P}ython.
\newblock \emph{Journal of Machine Learning Research}, 12:\penalty0 2825--2830,
  2011.

\bibitem[Rahmadi et~al.(2017)Rahmadi, Groot, Heins, Knoop, Heskes,
  et~al.]{rahmadi2017causality}
Ridho Rahmadi, Perry Groot, Marianne Heins, Hans Knoop, Tom Heskes, et~al.
\newblock Causality on cross-sectional data: Stable specification search in
  constrained structural equation modeling.
\newblock \emph{Applied Soft Computing}, 52:\penalty0 687--698, 2017.

\bibitem[Ren and Wang(2023)]{softmax_jingli}
Jingli Ren and Haiyan Wang.
\newblock Chapter 3 - calculus and optimization.
\newblock In Jingli Ren and Haiyan Wang, editors, \emph{Mathematical Methods in
  Data Science}, pages 51--89. Elsevier, 2023.
\newblock ISBN 978-0-443-18679-0.
\newblock \doi{https://doi.org/10.1016/B978-0-44-318679-0.00009-0}.
\newblock URL
  \url{https://www.sciencedirect.com/science/article/pii/B9780443186790000090}.

\bibitem[Ribeiro et~al.(2016)Ribeiro, Singh, and Guestrin]{ribeiro_why_2016}
Marco~Tulio Ribeiro, Sameer Singh, and Carlos Guestrin.
\newblock "{Why} {Should} {I} {Trust} {You}?": {Explaining} the {Predictions}
  of {Any} {Classifier}.
\newblock \emph{Proceedings of the 22nd ACM SIGKDD International Conference on
  Knowledge Discovery and Data Mining}, pages 1135--1144, August 2016.
\newblock \doi{10.1145/2939672.2939778}.
\newblock Conference Name: KDD '16: The 22nd ACM SIGKDD International
  Conference on Knowledge Discovery and Data Mining ISBN: 9781450342322 Place:
  San Francisco California USA Publisher: ACM.

\bibitem[Schermelleh-Engel et~al.(2003)Schermelleh-Engel, Moosbrugger, and
  M{\"u}ller]{schermelleh2003evaluating}
Karin Schermelleh-Engel, Helfried Moosbrugger, and Hans M{\"u}ller.
\newblock Evaluating the fit of structural equation models: Tests of
  significance and descriptive goodness-of-fit measures.
\newblock \emph{Methods of Psychological Research}, 2003.

\bibitem[Sch{\"o}lkopf et~al.(2021)Sch{\"o}lkopf, Locatello, Bauer, Ke,
  Kalchbrenner, Goyal, and Bengio]{scholkopf2021toward}
Bernhard Sch{\"o}lkopf, Francesco Locatello, Stefan Bauer, Nan~Rosemary Ke, Nal
  Kalchbrenner, Anirudh Goyal, and Yoshua Bengio.
\newblock Toward causal representation learning.
\newblock \emph{Proceedings of the IEEE}, 109\penalty0 (5):\penalty0 612--634,
  2021.

\bibitem[Sehwag et~al.(2022)Sehwag, Mahloujifar, Handina, Dai, Xiang, Chiang,
  and Mittal]{sehwag_robust_2022}
Vikash Sehwag, Saeed Mahloujifar, Tinashe Handina, Sihui Dai, Chong Xiang, Mung
  Chiang, and Prateek Mittal.
\newblock Robust {Learning} {Meets} {Generative} {Models}: can {Proxy}
  {Distributions} {Improve} {Adversarial} {Robustness}?
\newblock 2022.

\bibitem[Semenova et~al.(2022)Semenova, Rudin, and
  Parr]{semenova_existence_2022}
Lesia Semenova, Cynthia Rudin, and Ronald Parr.
\newblock On the {Existence} of {Simpler} {Machine} {Learning} {Models}.
\newblock In \emph{Proceedings of the 2022 {ACM} {Conference} on {Fairness},
  {Accountability}, and {Transparency}}, {FAccT} '22, pages 1827--1858, New
  York, NY, USA, June 2022. Association for Computing Machinery.
\newblock ISBN 978-1-4503-9352-2.
\newblock \doi{10.1145/3531146.3533232}.

\bibitem[Sharma and Kiciman(2020)]{sharma2020dowhyendtoendlibrarycausal}
Amit Sharma and Emre Kiciman.
\newblock Dowhy: An end-to-end library for causal inference, 2020.
\newblock URL \url{https://arxiv.org/abs/2011.04216}.

\bibitem[Shimizu et~al.(2006)Shimizu, Hoyer, Hyv\&\#228, rinen, and
  Kerminen]{shimizu_linear_2006}
Shohei Shimizu, Patrik~O. Hoyer, Aapo Hyv\&\#228, rinen, and Antti Kerminen.
\newblock A {Linear} {Non}-{Gaussian} {Acyclic} {Model} for {Causal}
  {Discovery}.
\newblock \emph{Journal of Machine Learning Research}, 7\penalty0
  (72):\penalty0 2003--2030, 2006.
\newblock ISSN 1533-7928.

\bibitem[Souiden et~al.(2022)Souiden, Omri, and Brahmi]{SOUIDEN2022100463}
Imen Souiden, Mohamed~Nazih Omri, and Zaki Brahmi.
\newblock A survey of outlier detection in high dimensional data streams.
\newblock \emph{Computer Science Review}, 44:\penalty0 100463, 2022.
\newblock ISSN 1574-0137.
\newblock \doi{https://doi.org/10.1016/j.cosrev.2022.100463}.
\newblock URL
  \url{https://www.sciencedirect.com/science/article/pii/S1574013722000107}.

\bibitem[Spirtes et~al.(2001)Spirtes, Glymour, and
  Scheines]{spirtes_causation_2001}
Peter Spirtes, Clark Glymour, and Richard Scheines.
\newblock \emph{Causation, {Prediction}, and {Search}}.
\newblock The MIT Press, January 2001.
\newblock ISBN 978-0-262-28415-8.
\newblock \doi{10.7551/mitpress/1754.001.0001}.

\bibitem[Sullivan et~al.(2021)Sullivan, Warkentin, and
  Wallace]{SULLIVAN2021530}
Joe~H. Sullivan, Merrill Warkentin, and Linda Wallace.
\newblock So many ways for assessing outliers: What really works and does it
  matter?
\newblock \emph{Journal of Business Research}, 132:\penalty0 530--543, 2021.
\newblock ISSN 0148-2963.

\bibitem[Tsymbal(2004)]{tsymbal2004problem}
Alexey Tsymbal.
\newblock The problem of concept drift: Definitions and related work.
\newblock \emph{Computer Science Department, Trinity College Dublin},
  106\penalty0 (2):\penalty0 58, 2004.

\bibitem[Ullman and Bentler(2012)]{ullman2012structural}
Jodie~B Ullman and Peter~M Bentler.
\newblock Structural equation modeling.
\newblock \emph{Handbook of psychology, second edition}, 2, 2012.

\bibitem[Van~Calster et~al.(2025)Van~Calster, Collins, Vickers, Wynants, Kerr,
  Barrene{\~n}ada, Varoquaux, Singh, Moons, Hernandez-Boussard, Timmerman,
  McLernon, van Smeden, and Steyerberg]{vancalster2025evaluation}
Ben Van~Calster, Gary~S. Collins, Andrew~J. Vickers, Laure Wynants, Kathleen~F.
  Kerr, Lasai Barrene{\~n}ada, Ga{\"e}l Varoquaux, Karandeep Singh, Karel G.~M.
  Moons, Tina Hernandez-Boussard, Dirk Timmerman, David~J. McLernon, Maarten
  van Smeden, and Ewout~W. Steyerberg.
\newblock Evaluation of performance measures in predictive artificial
  intelligence models to support medical decisions: Overview and guidance.
\newblock \emph{The Lancet Digital Health}, 7\penalty0 (12):\penalty0 e100916,
  2025.
\newblock \doi{10.1016/S2589-7500(25)00098-6}.

\bibitem[Vangala(2022)]{vangala2022mlops}
Vidyasagar Vangala.
\newblock Mlops in practice: A framework for scalable ai model deployment,
  monitoring, and retraining.
\newblock \emph{International Journal of Machine Learning Research in
  Cybersecurity and Artificial Intelligence}, 13\penalty0 (01):\penalty0
  740--753, 2022.

\bibitem[Wen et~al.(2021)Wen, Colon, Hansch, and Walsh]{wen2021causaltgan}
Bingyang Wen, Luis~Oala Colon, Ronny Hansch, and Brinnae Walsh.
\newblock Causal-{TGAN}: Generating tabular data using causal generative
  adversarial networks.
\newblock \emph{arXiv preprint arXiv:2104.10680}, 2021.

\bibitem[{Workplace Intelligence}(2020)]{oracle2020ai}
{Workplace Intelligence}.
\newblock As uncertainty remains, anxiety and stress reach a tipping point at
  work: {AI} at work 2020 study.
\newblock White paper, Oracle Corporation, October 2020.
\newblock URL
  \url{https://www.oracle.com/a/ocom/docs/applications/hcm/ai-at-work-2020.pdf}.

\bibitem[Xu et~al.(2019)Xu, Skoularidou, Cuesta-Infante, and
  Veeramachaneni]{xu2019modeling}
Lei Xu, Maria Skoularidou, Alfredo Cuesta-Infante, and Kalyan Veeramachaneni.
\newblock Modeling tabular data using conditional gan.
\newblock \emph{Advances in neural information processing systems}, 32, 2019.

\bibitem[Yong et~al.(2020)Yong, Fathy, and Brintrup]{yong2020bayesian}
Bang~Xiang Yong, Yasmin Fathy, and Alexandra Brintrup.
\newblock Bayesian autoencoders for drift detection in industrial environments.
\newblock In \emph{2020 IEEE international workshop on metrology for industry
  4.0 \& IoT}, pages 627--631. IEEE, 2020.

\bibitem[Yu et~al.(2025)Yu, Yuan, Wan, Tang, Kurdahi, and Liu]{yu2025addt}
Bo~Yu, Chaoran Yuan, Zishen Wan, Jie Tang, Fadi Kurdahi, and Shaoshan Liu.
\newblock Addt--a digital twin framework for proactive safety validation in
  autonomous driving systems.
\newblock \emph{arXiv preprint arXiv:2504.09461}, 2025.

\bibitem[Zhang et~al.(2023)Zhang, Wang, Fan, Cheung, and
  Bose]{zhang2023enhancing}
Xiaoge Zhang, Xiao-Lin Wang, Fenglei Fan, Yiu-Ming Cheung, and Indranil Bose.
\newblock Enhancing the performance of neural networks through causal discovery
  and integration of domain knowledge.
\newblock \emph{arXiv preprint arXiv:2311.17303}, 2023.

\bibitem[Zheng et~al.(2018)Zheng, Aragam, Ravikumar, and
  Xing]{zheng2018notears}
Xun Zheng, Bryon Aragam, Pradeep Ravikumar, and Eric~P. Xing.
\newblock {DAGs} with {NO} {TEARS}: Continuous optimization for structure
  learning.
\newblock In \emph{Advances in Neural Information Processing Systems}, 2018.

\bibitem[{\v{Z}}liobait{\.e}(2016)]{zliobaite2016overview}
Indr{\.e} {\v{Z}}liobait{\.e}.
\newblock An overview of concept drift applications.
\newblock In \emph{Big Data Analysis: New Algorithms for a New Society}, pages
  91--114. Springer, 2016.

\end{thebibliography}

\clearpage
\appendix
\section{OSMI Dataset Description}
\label{a:osmh-dataset}
Tables~\ref{tab:desc-dataset-1} and~\ref{tab:desc-dataset-2} list all 23 features of the OSMI dataset used in Section~\ref{ssb:datasets}, together with their type, possible values, and the corresponding survey question.

\begin{table}[H]
    \centering
    \begin{tabular}{p{3.6cm}c p{4.5cm} p{5cm}}
        \toprule
        \textbf{Name} & \textbf{Type} & \textbf{Possible Values} & \textbf{Description} \\
        \midrule
        \texttt{Age} & Numerical & Integer from 18 to 72 & Age of the Respondent\\
        \texttt{Gender} & Categorical & "Female", "Male", "Other" & Gender of the Respondent\\
        \texttt{self\_employed} & Categorical & "Yes", "No" & Are you self-employed?\\
        \texttt{family\_history} & Categorical & "Yes", 'No" & Do you have a family history of mental illness?\\
        \texttt{treatment} & Categorical & "Yes", "No" & Have you sought treatment for a mental health condition?\\
        \texttt{work\_interfere} & Categorical & "Often", "Sometimes",\newline"Rarely", "Never",\newline "No answer" & If you have a mental health condition, do you feel that it interferes with your work?\\
        \texttt{no\_employees} & Categorical & "1-5", "6-25", "26-100",\newline"100-500", "500-1000",\newline"More than 1000" & How many employees does your company or organization have? \\
        \texttt{remote\_work} & Categorical & "Yes", "No" & Do you work remotely at least 50\% of the time? \\
        \texttt{tech\_company} & Categorical & "Yes", "No" & Is your employer primarily a tech company/organization?\\
        \texttt{benefits} & Categorical & "Yes", "No", "Don't know" &Does your employer provide mental health benefits?\\
        \texttt{care\_options} & Categorical & "Yes", "No", "Not Sure" &Do you know the options for mental health care your employer provides?\\
        \texttt{wellness\_program} & Categorical & "Yes", "No", "Don't know" &Has your employer discussed mental health as part of an employee wellness program?\\
        \texttt{seek\_help} & Categorical & "Yes", "No", "Don't know" & Does your employer provide resources to learn about mental health and how to seek help?\\
        \texttt{anonymity} & Categorical & "Yes", "No", "Don't know" & Is your anonymity protected if you choose to take advantage of mental health or substance abuse treatment resources?\\
        \texttt{leave} & Categorical & "Very easy", "Somewhat easy", "Somewhat difficult", "Very difficult", "Don't know"& How easy is it for you to take medical leave for a mental health condition?\\
        \bottomrule
    \end{tabular}
    \caption{OSMI dataset overview, part 1.}
    \label{tab:desc-dataset-1}
\end{table}

\begin{table}[H]
    \centering
    \begin{tabular}{p{3.6cm}c p{4.5cm} p{5cm}}
        \toprule
        \textbf{Name} & \textbf{Type} & \textbf{Possible Values} & \textbf{Description} \\
        \midrule
         \texttt{mental\_health\newline\_consequence} & Categorical & "Yes", "Maybe", "No" & Do you think that discussing a mental health issue with your employer would have negative consequences?\\\texttt{phys\_health\newline\_consequence} & Categorical & "Yes", "Maybe", "No" & Do you think that discussing a physical health issue with your employer would have negative consequences?\\
        \texttt{coworkers} & Categorical & "Yes", "Some of them", "No" & Would you be willing to discuss a mental health issue with your coworkers?\\
        \texttt{supervisor} & Categorical & "Yes", "Some of them", "No" & Would you be willing to discuss a mental health issue with your direct supervisor(s)? \\
        \texttt{mental\_health\newline\_interview} & Categorical & "Yes", "Maybe", "No" & Would you bring up a mental health issue with a potential employer in an interview? \\
        \texttt{physical\_health\newline\_interview} & Categorical & "Yes", "Maybe", "No" & Would you bring up a physical  health issue with a potential employer in an interview? \\
        \texttt{mental\_vs\_physical} & Categorical & "Yes", "No","Don't Know" & Do you feel that your employer takes mental health as seriously as physical health?\\
        \texttt{obs\_consequence} & Categorical & "Yes", "No" & Have you heard of or observed negative consequences for coworkers with mental health conditions in your workplace?\\
        \bottomrule
    \end{tabular}
    \caption{OSMI dataset overview, part 2.}
    \label{tab:desc-dataset-2}
\end{table}

\clearpage
\section{Additional Sensitivity Analysis Figures}
\label{a:sensitivity-figures}
This appendix collects the topology and robustness-curve figures for Variants~B and~C of the sensitivity analysis (Section~\ref{ssb:sensitivity}). Variant~A is shown in the body as an illustrative example; the breaking points for all variants are summarized in Table~\ref{tab:sensitivity_table}.

\begin{figure}[H]
    \centering
    \includegraphics[width=1\linewidth]{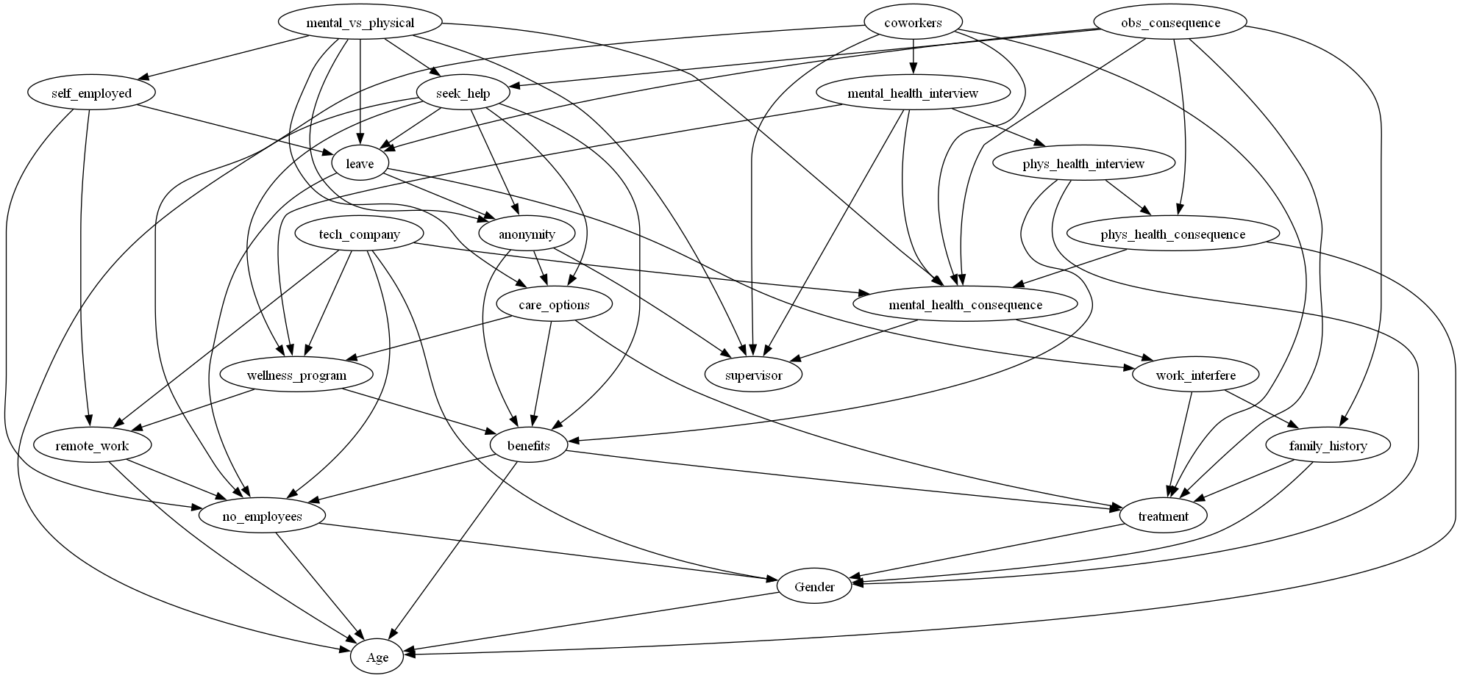}
    \caption{Variant~B: causal graph topology obtained with $\alpha_{PC}=0.2$ and no domain constraints.}
    \label{fig:osmh_topology_alpha-0.2}
\end{figure}
\begin{figure}[H]
    \centering
\includegraphics[width=1\linewidth]{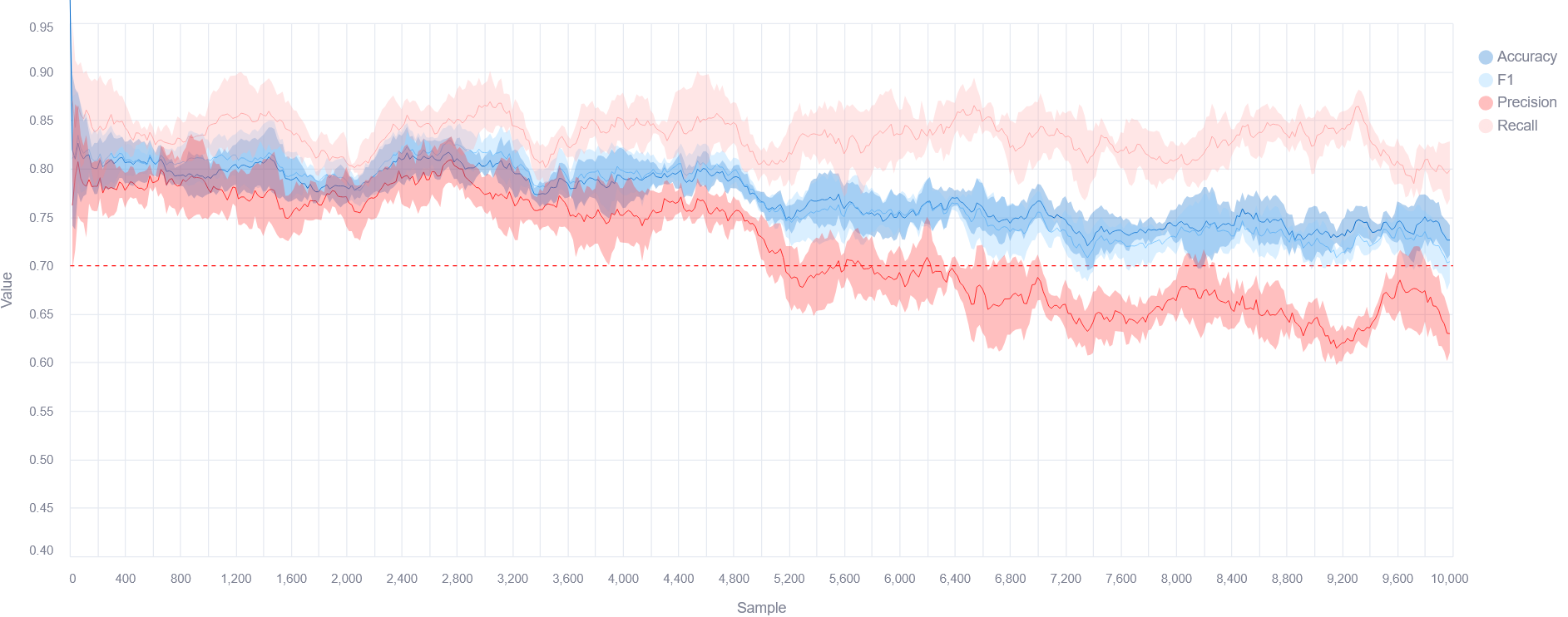}
\vspace{-0.55cm}
    \caption{XGBoost Robustness Curve under the Self Help scenario, using the Variant~B topology ($\alpha_{PC}=0.2$, no constraints). The red dashed line marks $\tau_{Precision}=0.7$.}
    \label{fig:robust-curve-alpha-0.2}
\end{figure}
\begin{figure}[H]
    \centering
\includegraphics[width=1\linewidth]{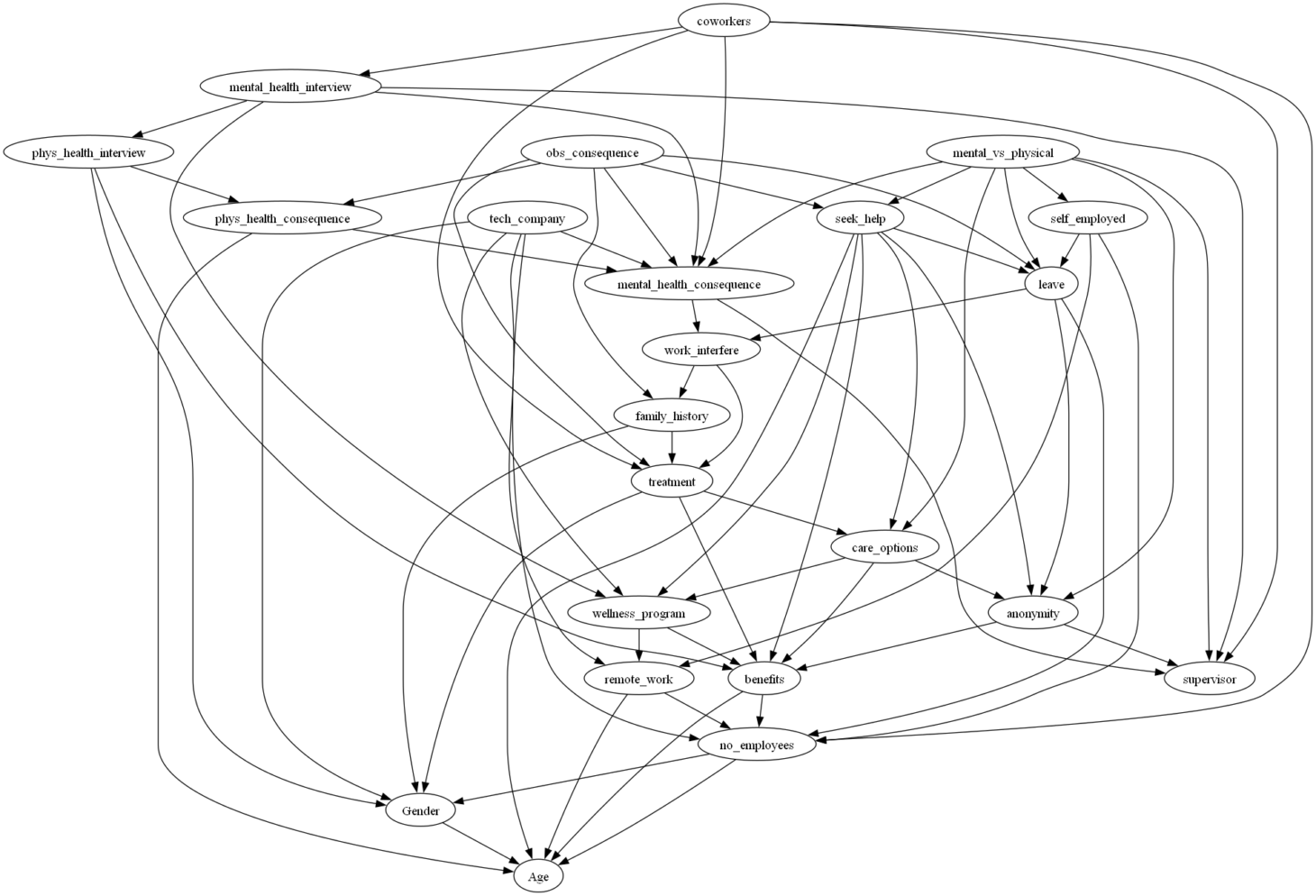}
    \caption{Variant~C: causal graph topology obtained with $\alpha_{PC}=0.2$ and an anti-causal domain constraint ($\texttt{treatment} \to \texttt{benefits}$).}
    \label{fig:osmh_topology_alpha-0.2_ac}
\end{figure}
\begin{figure}[H]
    \centering
\includegraphics[width=1\linewidth]{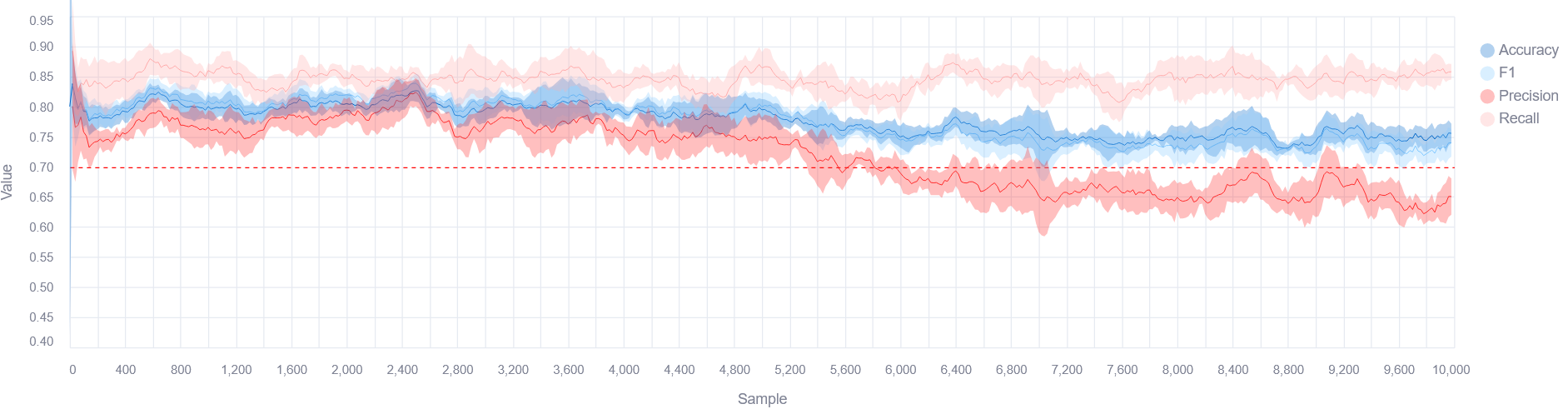}
\vspace{-0.55cm}
    \caption{XGBoost Robustness Curve under the Self Help scenario, using the anti-causal Variant~C topology. The Breaking Point shifts to $\approx -0.40$ but remains identifiable, supporting the robustness of the diagnostic.}
    \label{fig:anti-causal-robust-curve-alpha-0.2}
\end{figure}

\end{document}